\pdfoutput=1
\documentclass[twoside]{article}

\usepackage[accepted]{aistats2020}

\usepackage[round]{natbib}

\usepackage[utf8]{inputenc} 
\usepackage[T1]{fontenc}    
\usepackage[colorlinks=true,allcolors=black]{hyperref} 
\usepackage{url}            
\usepackage{booktabs}       
\usepackage{amsfonts}       
\usepackage{nicefrac}       
\usepackage{microtype}

\usepackage{todonotes}

\usepackage{amsmath, amsfonts, amssymb, stmaryrd}
\usepackage{caption}
\usepackage{blindtext}
\usepackage{ifthen}

\usepackage{xcolor}
\usepackage{tikz}
\usepackage{tikz-3dplot}
\usepackage[customcolors]{hf-tikz}
\usetikzlibrary{shapes,
                shadings,
                calc,
                arrows,
                backgrounds,
                colorbrewer,
                shadows.blur}
\usepackage{ifthen}
\usepackage{comment}
 \usepackage{pdfpages}

\definecolor{YlGnBu-5-1}{RGB}{255,255,204}
\definecolor{YlGnBu-5-2}{RGB}{161,218,180}
\definecolor{YlGnBu-5-3}{RGB}{65,182,196}
\definecolor{YlGnBu-5-4}{RGB}{44,127,184}
\definecolor{YlGnBu-5-5}{RGB}{37,52,148}
\definecolor{RdPu-5-1}{RGB}{254,235,226}
\definecolor{RdPu-5-2}{RGB}{251,180,185}
\definecolor{RdPu-5-3}{RGB}{247,104,161}
\definecolor{RdPu-5-4}{RGB}{197,27,138}
\definecolor{RdPu-5-5}{RGB}{122,1,119}

\colorlet{forwardFill}{YlGnBu-5-1}
\colorlet{forwardDraw}{black}
\colorlet{backwardGradientFill}{YlGnBu-5-2}
\colorlet{backwardGradientDraw}{black}
\colorlet{backwardHessianFill}{YlGnBu-5-3}
\colorlet{backwardHessianDraw}{black}

\colorlet{thickArrow}{black}
\colorlet{thickParamArrow}{black}

\colorlet{moduleFill}{black!50!white}
\colorlet{moduleFrameFill}{moduleFill!50!white}

\colorlet{moduleDraw}{black}
\colorlet{moduleFrameDraw}{moduleDraw}

\pgfmathsetmacro{\nodeMinWidth}{7.5}
\pgfmathsetmacro{\nodeMinHeight}{4}
\pgfmathsetmacro{\vNodeDistance}{4.5}
\pgfmathsetmacro{\hNodeDistance}{12.5}

\pgfmathsetmacro{\moduleMinHeight}{3*\vNodeDistance}

\pgfmathsetmacro{\paramArrowYShift}{-0.7}
\pgfmathsetmacro{\paramArrowSplit}{2}

\tikzset{thickArrow/.style={
		->,
		>=stealth,
		thick,
		draw=thickArrow,
		rounded corners=1ex
	}
}

\tikzset{thickParamArrow/.style={
		thickArrow,
		draw=thickParamArrow,
	}
}

\tikzset{moduleFrame/.style={
		rounded corners=1ex,
		minimum height=\moduleMinHeight ex,
		draw=moduleFrameDraw,
		fill=moduleFrameFill,
		inner sep=0
	}
}

\tikzset{module/.style={
		rectangle,
		rounded corners=1ex,
		minimum height=\moduleMinHeight ex,
		draw=moduleDraw,
		fill=moduleFill,
		inner sep=0
	}
}

\tikzset{forward/.style={
		rectangle,
		rounded corners=1ex,
		minimum width=\nodeMinWidth ex,
		minimum height=\nodeMinHeight ex,
		draw=forwardDraw,
		fill=forwardFill
	}
}

\tikzset{backwardGradient/.style={
		rectangle,
		rounded corners=1ex,
		minimum width=\nodeMinWidth ex,
		minimum height=\nodeMinHeight ex,
		draw=backwardGradientDraw,
		fill=backwardGradientFill
	}
}

\tikzset{backwardHessian/.style={
		rectangle,
		rounded corners=1ex,
		minimum width=\nodeMinWidth ex,
		minimum height=\nodeMinHeight ex,
		draw=backwardHessianDraw,
		fill=backwardHessianFill
	}
}

\newcommand{\drawMessages}[3]{
	\ifthenelse{\equal{#1}{ }}{}{
        \node (input)
            [forward]
            { #1 };
	}
	\ifthenelse{\equal{#2}{ }}{}{	
	\node (inputGradient)
		[backwardGradient,
		below of=input,
		node distance=\vNodeDistance ex]
		{ #2 };	
	}
	\ifthenelse{\equal{#3}{ }}{
        \phantom{
            \node (inputHessian)
            [backwardHessian,
            below of=input,
            node distance=2*\vNodeDistance ex]
            { #3 };		
        }	
	}{	
        \node (inputHessian)
            [backwardHessian,
            below of=input,
            node distance=2*\vNodeDistance ex]
            { #3 };
	}	
}

\newcommand{\drawMessagesWithArrows}[4]{
	\drawMessages{#1}{#2}{#3}
	\begin{pgfonlayer}{background}
        \ifthenelse{\equal{#1}{ }}{}{
            \draw [thickArrow]
                ($(input)+(-{#4 ex/2},0)$) to ++(#4 ex, 0);
        }
        \ifthenelse{\equal{#2}{ }}{}{
            \draw [thickArrow, <-]
                ($(inputGradient)+(-{#4 ex/2},0)$) to ++(#4 ex, 0);
        }
        \ifthenelse{\equal{#3}{ }}{}{
            \draw [thickArrow, <-]
                ($(inputHessian)+(-{#4 ex/2},0)$) to ++(#4 ex, 0);
        }	
	\end{pgfonlayer}
}

\newcommand{\drawParamsWithArrows}[3]{
	\drawMessages{#1}{#2}{#3}
	\ifthenelse{\equal{#1}{ }}{}{
		\coordinate (arrowParam) at
             ($(inputHessian.south east)+(3*\paramArrowSplit ex, \paramArrowYShift ex)$);
		\draw [thickParamArrow] (input.east) -| (arrowParam);
	}
	\ifthenelse{\equal{#2}{ }}{}{
		\coordinate (arrowGradient) at
             ($(inputHessian.south east)+(2*\paramArrowSplit ex, \paramArrowYShift ex)$);
		\draw [thickParamArrow] (arrowGradient) |- (inputGradient.east);
	}
	\ifthenelse{\equal{#3}{ }}{}{
		\coordinate (arrowHessian) at
            ($(inputHessian.south east)+(\paramArrowSplit ex, \paramArrowYShift ex)$);
		\draw [thickParamArrow] (arrowHessian) |- (inputHessian.east);
	}	
}

\newcommand{\drawModuleWithParams}[5]{			
	\node (moduleFrame) [moduleFrame] {
		\tikz{
			\node (module)
                [module, minimum width=#2 ex]
                { #1 };
			\node (moduleParams)
                [anchor=south]
                at (module.north)
                {\tikz \drawParamsWithArrows{#3}{#4}{#5};};
		}
	};			
}

\newcommand{\drawModuleNoParams}[2]{			
	\node (module) [module, minimum width=#2 ex] { #1 };			
}

\newcommand{\drawGrid}[5]{
  \foreach \i in {1, ..., #2}{
    \foreach \j in {#3, ..., 1}{
      \draw[canvas is yz plane at x=#1,
            transform shape,
            draw=black,
            fill=#4,
            drop shadow=black]
      (\i,\j) rectangle ++(-1, -1) coordinate [pos=0.5] (#5-\i-\j);
    }
}
}

\newcommand{\drawGridNoShadow}[7]{
  \foreach \i in {#6, ..., #2}{
    \foreach \j in {#3, ..., #7}{
      \draw[canvas is yz plane at x=#1,
            transform shape,
            draw=black,
            fill=#4]
      (\i,\j) rectangle ++(-1, -1) coordinate [pos=0.5] (#5-\i-\j);
    }
}
}

\newcommand{\drawGridTextLabel}[4]{
  \node[canvas is yz plane at x=#1,
        transform shape] at ($(0, -0.5,-0.5) + (0, #2, #3)$) {#4};
}

\colorlet{foregroundChannel}{YlGnBu-5-2}
\colorlet{centerChannel}{YlGnBu-5-3}
\colorlet{backgroundChannel}{YlGnBu-5-4}
\colorlet{foregroundOutput}{RdPu-5-2}
\colorlet{backgroundOutput}{RdPu-5-4}

\tikzset{style inputBackground/.style={
    set fill color=backgroundChannel,
    set border color=black,
    thin,
    rounded corners=0pt
  },
  style inputCenter/.style={
    set fill color=centerChannel,
    set border color=black,
    thin,
    rounded corners=0pt
  },
  style inputForeground/.style={
    set fill color=foregroundChannel,
    set border color=black,
    thin,
    rounded corners=0pt
  },
  style inputBackgroundLight/.style={
    set fill color=backgroundChannel!50!white,
    set border color=black,
    thin,
    rounded corners=0pt
  },
  style inputCenterLight/.style={
    set fill color=centerChannel!50!white,
    set border color=black,
    thin,
    rounded corners=0pt
  },
  style inputForegroundLight/.style={
    set fill color=foregroundChannel!50!white,
    set border color=black,
    thin,
    rounded corners=0pt
  },
  horizontal/.style={
    above left offset={-0.15,0.31},
    below right offset={0.15,-0.125},
    #1
  },
  vertical/.style={
    above left offset={-0.1,0.3},
    below right offset={0.15,-0.15},
    #1
  },
  style outputForeground/.style={
    set fill color=foregroundOutput,
    set border color=black,
    thin,
    rounded corners=0pt
  },
   style outputBackground/.style={
    set fill color=backgroundOutput,
    set border color=black,
    thin,
    rounded corners=0pt
  },
   style outputForegroundLight/.style={
    set fill color=foregroundOutput!50!white,
    set border color=black,
    thin,
    rounded corners=0pt
  },
   style outputBackgroundLight/.style={
    set fill color=backgroundOutput!50!white,
    set border color=black,
    thin,
    rounded corners=0pt
  },
}

 \usepackage{pgfplots}

\newcommand{\tensor}[1]{\ensuremath{\mathsf{#1}}}
\renewcommand{\vec}{\operatorname{vec}}
\newcommand{\diag}{\operatorname{diag}}
\newcommand{\D}{\mathrm{D}}
\newcommand{\G}{\mathrm{G}}
\newcommand{\F}{\mathrm{F}}
\newcommand{\He}{\mathrm{H}}
\newcommand{\HeCal}{\mathcal{H}}
\newcommand{\E}{\operatorname{\mathbb{E}}}
\newcommand{\diff}{\mathrm{d}}
\newcommand{\average}[1]{\overline{ #1 }}

\newlength{\figwidth}
\newlength{\figheight}
\pgfkeys{/pgfplots/original/.style={
        enlarge x limits = 0.,
        enlarge y limits = 0.02,
        width=\figwidth,
        height=\figheight,
        grid=major,
        grid style = {dashed},
        every axis plot/.append style={line width = 1.5pt},
        tick pos = left,
        xtick align = inside,
        ytick align = inside,
        legend cell align = left,
        legend style = {fill opacity = 0.7,
            text opacity = 1}
    }}
    
\newcommand{\resetPGFStyle}{
    \pgfkeys{/pgfplots/mystyle/.style={
            original
    }}
}
\resetPGFStyle
\usepackage{nameref, zref-xr}
\zxrsetup{toltxlabel}
\zexternaldocument*{supplementary}

\newcommand\blankFootnote[1]{
	\begingroup
	\renewcommand\thefootnote{}\footnote{#1}
	\addtocounter{footnote}{-1}
	\endgroup
}

\begin{document}

\runningtitle{Modular Block-diagonal Curvature Approximations for 
	Feedforward Architectures}

\twocolumn[

\aistatstitle{Modular Block-diagonal Curvature Approximations\newline for 
Feedforward Architectures}

 \aistatsauthor{ Felix Dangel \And  Stefan Harmeling \And  Philipp Hennig }

 \aistatsaddress{ University of T\"ubingen \\ \texttt{fdangel@tue.mpg.de}
   \And
   Heinrich Heine University\\ D\"usseldorf \\ \texttt{harmeling@hhu.de}
   \And
   University of T\"ubingen and\\ MPI for Intelligent Systems, T\"ubingen \\ \texttt{ph@tue.mpg.de}
}

]

\begin{abstract}
    We propose a modular extension of backpropagation for the computation of 
    block-diagonal approximations to various curvature matrices of the 
    training objective (in particular, the Hessian, generalized Gauss-Newton, 
    and positive-curvature Hessian). The approach reduces the otherwise 
    tedious manual derivation of these matrices into local modules, and 
    is easy to integrate into existing machine learning libraries.
    Moreover, we develop a compact notation derived from matrix 
    differential calculus. We outline different strategies applicable to our 
    method. They subsume recently-proposed block-diagonal approximations as 
    special cases, and are extended to 
    convolutional neural networks in this work.
\end{abstract}

\section{Introduction}
Gradient backpropagation is the central computational operation of 
contemporary deep learning. Its modular structure allows easy extension 
across network architectures, and thus automatic computation of gradients 
given the computational graph of the forward pass \citep[for a review, 
see][]{baydin2018Autodiff}. But optimization using only the first-order information of the objective's gradient can be unstable and slow, due to ``vanishing'' or ``exploding'' behaviour of the gradient. 
Incorporating curvature, second-order methods can avoid such 
scaling issues and converge in fewer iterations. 
Such methods locally approximate the objective function $E$ by a quadratic 
\begin{math}
    E(x) + \delta x^\top (x_* - x) + \frac{1}{2} (x_* - x)^\top C (x_* 
    - x)
\end{math}
around the current location $x$, 
using the gradient $\delta x = \nicefrac{\partial E}{\partial x}$ and a 
positive 
semi-definite (PSD) curvature matrix $C$ --- the Hessian 
of $E$ or approximations thereof. The quadratic is minimized by 
\begin{align}
    \label{equ:NewtonUpdate}
    x_* = x 
+ \Delta x \quad \text{with} \quad \Delta x = -C^{-1} \delta x\,.
\end{align}
Computing the update step requires that the $C\Delta x = -\delta x$ linear system be solved. To accomplish this task, providing a matrix-vector multiplication with the curvature matrix $C$ is sufficient.\blankFootnote{Code available at \href{https://github.com/f-dangel/hbp}{\texttt{github.com/f-dangel/hbp}}.}

\paragraph{Approaches to second-order optimization:}

For some curvature matrices, exact multiplication can be performed at the 
cost of one backward pass by automatic differentiation 
\citep{pearlmutter1994FastExactHessianMultiplication, 
schraudolph2002FastMatrixVectorProducts}. This \emph{matrix-free} 
formulation can then be leveraged to solve~\eqref{equ:NewtonUpdate} using iterative solvers such as the method 
of conjugate gradients~(CG)~\citep{martens2010HessianFree}. However, since 
this linear solver can still 
require multiple iterations,
the 
increased per-iteration progress of the resulting optimizer might be 
compensated by increased computational cost. Recently, a parallel version of 
Hessian-free optimization was proposed 
in~\citep{zhang2017BlockDiagonalHessianFree},
which only considers the content 
of Hessian sub-blocks along the diagonal. Reducing the Hessian to a 
block diagonal allows for parallelization, tends to lower the 
required number of CG iterations, and seems to improve the optimizer's performance. 

    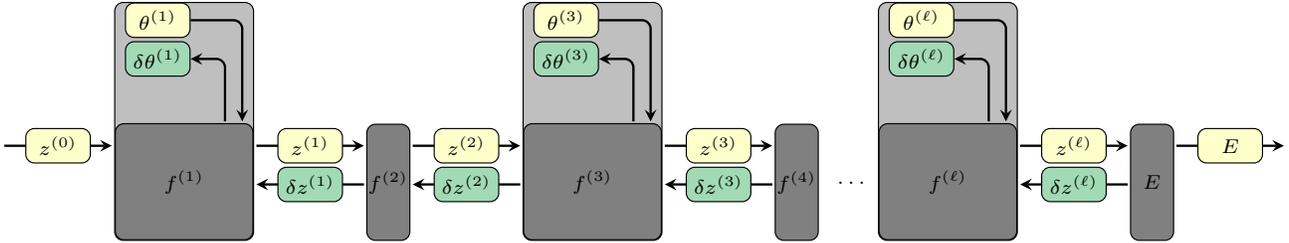
\begin{figure*}[t]
		\centering
		\resizebox{\linewidth}{!}
		{\scriptsize
			
\begin{tikzpicture}
    \node (in1) 
        [inner sep=0]
        {\tikz \drawMessagesWithArrows{$z^{(0)}$}{ }{ }{\hNodeDistance};};
    \node (layer1)
        [anchor=south west, inner sep=0]
        at (in1.south east)
        {\tikz \drawModuleWithParams{$f^{(1)}$}{16}{$\theta^{(1)}$}{$\delta \theta^{(1)}$}{ };};
    \node (out1)
        [inner sep=0, anchor=south west]
        at (layer1.south east)
        {\tikz \drawMessagesWithArrows{$z^{(1)}$}{$\delta z^{(1)}$}{ }{\hNodeDistance};};
    \node (layer2) 
        [inner sep=0pt, anchor=south west]
        at (out1.south east)
        {\tikz\drawModuleNoParams{$f^{(2)}$}{5};};
    
    \node (in2)
        [inner sep=0, anchor=south west]
        at (layer2.south east)
        {\tikz \drawMessagesWithArrows{$z^{(2)}$}{$\delta z^{(2)}$}{ }{\hNodeDistance};};
    \node (layer3)
        [anchor=south west, inner sep=0]
        at (in2.south east)
        {\tikz \drawModuleWithParams{$f^{(3)}$}{16}{$\theta^{(3)}$}{$\delta \theta^{(3)}$}{ };};
    \node (out3)
        [inner sep=0, anchor=south west]
        at (layer3.south east)
        {\tikz \drawMessagesWithArrows{$z^{(3)}$}{$\delta z^{(3)}$}{ }{\hNodeDistance};};
    \node (layer4)
        [inner sep=0pt, anchor=south west]
        at (out3.south east)
        {\tikz\drawModuleNoParams{$f^{(4)}$}{5};};	

    \node (dots)
        [xshift=2ex, inner sep=0pt, anchor=west]
        at (layer4.east)
        {$\dots$};
    \node (layer5)
        [xshift=12ex, anchor=south west, inner sep=0]
        at (out3.south east)
        {\tikz \drawModuleWithParams{$f^{(\ell)}$}{16}{$\theta^{(\ell)}$}{$\delta \theta^{(\ell)}$}{ };};
    \node (out4)
        [inner sep=0, anchor=south west]
        at (layer5.south east)
        {\tikz \drawMessagesWithArrows{$z^{(\ell)}$}{$\delta z^{(\ell)}$}{ }{\hNodeDistance};};
    
    \node (lossLayer)
        [inner sep=0pt, anchor=south west]
        at (out4.south east)
        {\tikz\drawModuleNoParams{$E$}{5};};
    \node (loss)
        [inner sep=0, anchor=south west]
        at (lossLayer.south east)
        {\tikz \drawMessagesWithArrows{$E$}{ }{ }{\hNodeDistance};};	
\end{tikzpicture} 		}
		\caption{Standard feedforward network architecture, i.e.~the 
			repetition of affine transformations parameterized by $ 
			\theta^{(i)} = (W^{(i)}, b^{(i)})$ followed by elementwise 
			activations. Arrows from left to right and vice versa 
			indicate the data flow during forward pass and gradient 
			backpropagation, respectively.}
		\label{fig:setting}
	\end{figure*}

There have also been attempts to compute parts of the Hessian in an iterative 
fashion \citep{mizutani2008StagewiseSecondOrderBackpropagation}.
Storing these constituents efficiently often requires an involved manual 
analysis of the Hessian's structure, leveraging its outer-product form in 
many scenarios \citep{naumov2017HessianInMatrixForm, 
bakker2018OuterProductStructure}. Recent works developed different 
block-diagonal approximations (BDA) of curvature matrices that provide fast 
multiplication~\citep{martens2015KFAC, grosse2016KFACConvolution, 
botev2017PracticalGaussNewton, chen2018BDAPCH}.

These works have repeatedly shown that, empirically, second-order information can
improve the training of deep learning problems. Perhaps the most important 
practical hurdle to the adoption of second-order optimizers is that 
they tend to be tedious to integrate in existing machine learning frameworks 
because they
require manual implementations. As efficient automated 
implementations have arguably been more important for the wide-spread use of 
deep learning than many conceptual advances, we
aim to develop a framework that makes computation of Hessian approximations 
about as easy and automated as gradient backpropagation.

\paragraph{Contribution:} 

This paper introduces a modular formalism for the 
computation of block-diagonal approximations of Hessian and curvature 
matrices, to various block resolutions, for feedforward neural networks. The 
framework unifies previous approaches in a form that, similar to 
gradient backpropagation, reduces implementation and analysis to local 
modules. 
Following the design pattern of gradient backprop also has the 
advantage that this formalism can readily be integrated into 
existing machine learning libraries, and flexibly modified for different 
block groupings and approximations. 

The framework consists of three principal parts:
\begin{enumerate}
	\item a modular formulation for \emph{exact} computation of Hessian block 
	diagonals of feedforward neural nets.
    We achieve a clear presentation 
    by leveraging the notation of matrix differential 
    calculus~\citep{magnus1999MatrixDifferentialCalculus}.
	\item projections
    onto the positive semi-definite cone
    by eliminating sources of concavity.
	\item backpropagation strategies to obtain (i) exact curvature 
	matrix-vector products (with previously inaccessible BDAs of 
	the Hessian) and 
	(ii) further approximated multiplication routines that save computations 
	by evaluating the matrix representations of intermediate quantities once, 
	at the cost of additional memory consumption.
\end{enumerate}
The first two contributions can be understood as an explicit formulation of 
well-known tricks for fast multiplication by curvature matrices using 
automatic 
differentiation~\citep{pearlmutter1994FastExactHessianMultiplication, 
schraudolph2002FastMatrixVectorProducts}.
However, we also address a new class of curvature matrices, the 
positive-curvature Hessian (PCH) introduced in~\cite{chen2018BDAPCH}.
Our solutions to the latter two points are generalizations of previous 
works~\citep{botev2017PracticalGaussNewton,chen2018BDAPCH} to 
the 
fully modular case, which become accessible due to the first contribution. 
They represent additional modifications to  make the scheme 
computationally tractable and obtain curvature approximations with 
desirable properties for 
optimization.

\section{Notation}
\label{sec:basicSetting}

    We consider feedforward neural networks composed of
    $\ell$ modules $f^{(i)}, i = 1, \ldots, \ell$, which can be represented 
    as a
    computational graph mapping the input $z^{(0)}=x$ to the 
    output $z^{(\ell)}$ (Figure~\ref{fig:setting}).
    A module~$f^{(i)}$ receives the parental output~$z^{(i-1)}$, applies an
    operation involving the network parameters~$\theta^{(i)}$, and sends the 
    output 
    $z^{(i)}$ to its 
    child. Thus, $f^{(i)}$ is of the form
    \begin{math}
        z^{(i)} = f^{(i)}(z^{(i-1)}, \theta^{(i)}). 
    \end{math}
    Typical choices include elementwise nonlinear
    activation without any parameters and affine transformations 
    $z^{(i)} = W^{(i)} z^{(i-1)} + b^{(i)}$ with parameters given by 
    the weights $W^{(i)}$ and the
    bias $b^{(i)} $. Affine and activation modules are usually 
    considered as a single conceptual unit, one \emph{layer} of the network.
    However, for backpropagation of derivatives it is simpler to consider 
    them separately as two \emph{modules}. 

    \begin{figure*}[t]
		\begin{minipage}{0.45\linewidth}
			\centering
			\resizebox{!}{2.6cm}{
				{
					\begin{tikzpicture}
    \node (in)
        [inner sep=0]
        {\tikz \drawMessagesWithArrows{$x$}{$\delta x$}{$\HeCal x$}{\hNodeDistance};};
    \node (module)
         [anchor=south west, inner sep=0]
         at (in.south east)
         {\tikz \drawModuleWithParams{$f$}{16}{$\theta$}{$\delta\theta$}{$\HeCal \theta$};};
    \node (out)
        [inner sep=0, anchor=south west]
        at (module.south east)
         {\tikz \drawMessagesWithArrows{$z$}{$\delta z$}{$\HeCal z$}{\hNodeDistance};};	
\end{tikzpicture} 				}
			}
		\end{minipage}
		\hfill
		\begin{minipage}{0.5\linewidth}
			\caption{Forward pass, gradient backpropagation, and Hessian 
				backpropagation for a single module. 
				Arrows from left to right 
				indicate the data flow in the forward pass $z = f(x, 
				\theta)$, 
				while the opposite 
				orientation indicates the gradient backpropagation by 
				Equation~\eqref{equ:gradientBackpropagation}. We suggest to 
				extend this by the backpropagation of the Hessian as 
				indicated by 
				Equation~\eqref{equ:hessianBackPropagation}.}
			\label{fig:sketchModule}
		\end{minipage}
	\end{figure*}

    Given the network output $z^{(\ell)}(x, \theta^{(1, \dots, \ell)})$ 
    of a datum $x$ with label $y$, the goal is to minimize the expected 
    risk of the loss function $E(z^{(\ell)}, y)$.
    Under the framework of empirical risk minimization,
    the parameters 
    are tuned to optimize the loss on the training set $Q= 
    \left\{(x,y)_{i=1}^N\right\}$, 
    \begin{align}
        \label{equ:objective}
        \min_{\theta^{(1,\dots, \ell)}} \frac{1}{|Q|} \sum_{(x, y) \in Q} 
        E(z^{(\ell)}(x), y)\,. 
    \end{align}
    In practice, the objective is typically further approximated 
    stochastically by drawing a mini-batch $B \subset Q$ from 
    the training set. We will treat both scenarios without further 
    distinction, since the structure relevant to our purposes is that 
    Equation~\eqref{equ:objective} is an 
    average of terms depending on individual data points.
    Quantities for optimization, be it gradients or second 
    derivatives of the loss with respect to the network parameters, can be 
    processed in parallel, then averaged.

\section{Main contribution}
\label{sec:modularApproach}
    First-order auto-differentiation for a custom module requires the 
    definition of only two local operations, \emph{forward} and 
    \emph{backward}, 
    whose outputs are propagated along the computation graph. This 
    modularity facilitates the extension of gradient backpropagation 
    by new operations, which can then be used to build networks by 
    composition. To illustrate the principle, we consider a single 
    module from the network of Figure~\ref{fig:setting}, depicted in 
    Figure~\ref{fig:sketchModule}, in this section. The forward pass $f(x,\theta)$ maps
    the input $x$ to the output $z$ by means of the module 
    parameters $\theta$ (to simplify notation, we drop layer indices). All 
    quantities are assumed to be vector-shaped (tensor-valued 
    quantities can be vectorized, see 
    Section~\ref{sec:matrixDifferentialCalculus} 
    of the Supplements). Optimization requires the gradient of the
    loss function with respect to the parameters, 
    $\nicefrac{\partial E(\theta)}{\partial \theta} = \delta\theta$.
    We will use the shorthand 
    \begin{align}
    	\delta\cdot = \frac{\partial E(\cdot)}{\partial \vec(\cdot)}\,.
    \end{align}
    During gradient backpropagation the module receives the loss gradient 
    with respect to its output, $\delta z$, from its child. The backward  
    operation computes gradients with respect to the module 
    parameters and input, $\delta \theta$ and $\delta x$ from 
    $\delta z$. Backpropagation continues by sending the gradient with 
    respect to the module's input to its parent, which proceeds in the same 
    way (see Figure~\ref{fig:setting}).
    By the chain rule, gradients with respect to an element of the module's 
    input can be computed as $\delta x_i = \sum_j (\nicefrac{\partial 
    z_j}{\partial x_i}) \delta z_j$. The vectorized version is compactly 
    written in terms of the Jacobian matrix $\D z(x) = \nicefrac{\partial 
    z(x)}{\partial x^\top}$, which contains all partial derivatives of $z$ 
    with respect to $x$. The arrangement of partial derivatives is such that 
$[\D z(x)]_{j,i} = \nicefrac{\partial z_j(x)}{\partial x_i}$, i.e.
    \begin{align}
        \delta x &= \left[\D z(x)\right]^\top \delta z\,.
        \label{equ:gradientBackpropagation}
    \end{align}
    Analogously, the parameter gradients are given by 
    $\delta\theta_i = \sum_j \frac{\partial z_j}{\partial \theta_i} \delta 
    z_j$,
    i.e.\ $\delta\theta = \left[ \D z(\theta) \right]^\top \delta z$,
    which reflects the symmetry of both $x$ and $\theta$ acting
    as input to the module. Implementing gradient backpropagation
    thus requires multiplications by (transposed) Jacobians. 

    We can apply the chain rule a second time to obtain expressions for second-order partial derivatives of the loss function $E$
    with respect to elements of $x$ or $\theta$, 
    \begin{align}
    	\label{equ:chainRuleComponentwise}
		\begin{split}
	       \!\frac{\partial^2 E(x)}{\partial x_i \partial x_j} &= 
	       \frac{\partial}{\partial x_j}\left( \sum_k \frac{\partial 
	       z_k}{\partial x_i} \delta z_k \right)
           \\
           &= \sum_{k, l} \frac{\partial z_k}{\partial x_i} \frac{\partial^2 
	       E(z)}{\partial z_k \partial z_l}	       
	       \frac{\partial z_l}{\partial x_j} +\sum_k \frac{\partial^2 z_k}{\partial x_i \partial x_j} \delta z_k\,,
		\end{split}
    \end{align}
    by means of $\nicefrac{\partial}{\partial x_j} = \sum_l 
    (\nicefrac{\partial z_l}{\partial x_j}) \nicefrac{\partial}{\partial 
    z_l}$ and the product rule.
    The first term of Equation~\eqref{equ:chainRuleComponentwise} propagates 
    curvature information of the output further back, while the second 
    term introduces second-order effects of
    the module itself. Using the Hessian matrix $\He E(x) = 
    \nicefrac{\partial^2 E(x)}{(\partial x^\top \partial x)}$ of a scalar 
    function with respect to a vector-shaped quantity $x$, the 
    Hessian of the loss function will be abbreviated by
    \begin{align}
    	\He E(\cdot) = \HeCal \cdot = \frac{\partial^2 
    	E(\cdot)}{\partial 
    	\vec(\cdot)^\top \partial \vec(\cdot)}\,,
    \end{align}
    which results in the matrix version of Equation~\eqref{equ:chainRuleComponentwise},
    \begin{align}
    	\label{equ:hessianBackPropagation}
    	\HeCal x &= \left[\D z(x)\right]^\top \HeCal z \left[\D z(x)\right] + 
    	\sum_k \left[\He z_k(x)\right] \delta z_k\,.
    \end{align}
    Note that the second-order effect introduced by the module itself via 
    $\He z_k(x)$ vanishes if $f_k(x, \theta)$ is linear in
    $x$. Because the layer parameters $\theta$ can be regarded as inputs to 
    the layer, they are treated in exactly the same way, replacing $x$ by 
    $\theta$ in the above expression.

    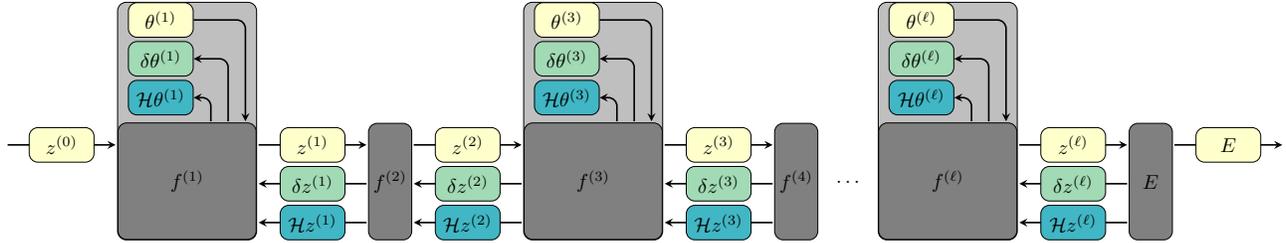
\begin{figure*}[t]
		\centering
		\resizebox{\linewidth}{!}
		{
			
\begin{tikzpicture}
    \node (in1)
        [inner sep=0]
        {\tikz \drawMessagesWithArrows{$z^{(0)}$}{ }{ }{\hNodeDistance};};
    \node (layer1)
        [anchor=south west, inner sep=0]
        at (in1.south east)
        {\tikz \drawModuleWithParams{$f^{(1)}$}{16}{$\theta^{(1)}$}{$\delta \theta^{(1)}$}{$\HeCal \theta^{(1)}$};};
    \node (out1)
        [inner sep=0, anchor=south west]
        at (layer1.south east)
         {\tikz \drawMessagesWithArrows{$z^{(1)}$}{$\delta  z^{(1)}$}{$\HeCal  z^{(1)}$}{\hNodeDistance};};
    \node (layer2)
        [inner sep=0pt, anchor=south west]
        at (out1.south east)
        {\tikz\drawModuleNoParams{$f^{(2)}$}{5};};
    \node (in2)
        [inner sep=0, anchor=south west]
        at (layer2.south east)
        {\tikz \drawMessagesWithArrows{$z^{(2)}$}{$\delta z^{(2)}$}{$\HeCal  z^{(2)}$}{\hNodeDistance};};
    \node (layer3)
        [anchor=south west, inner sep=0]
        at (in2.south east)
        {\tikz \drawModuleWithParams{$f^{(3)}$}{16}{$\theta^{(3)}$}{$\delta \theta^{(3)}$}{$\HeCal \theta^{(3)}$};};
    \node (out3)
        [inner sep=0, anchor=south west]
        at (layer3.south east)
        {\tikz \drawMessagesWithArrows{$z^{(3)}$}{$\delta  z^{(3)}$}{$\HeCal  z^{(3)}$}{\hNodeDistance};};
    \node (layer4)
        [inner sep=0pt, anchor=south west]
        at (out3.south east)
        {\tikz\drawModuleNoParams{$f^{(4)}$}{5};};	
    \node (dots)
        [xshift=2ex, inner sep=0pt, anchor=west]
         at (layer4.east) {$\dots$};

    \node (layer5)
        [xshift=12ex, anchor=south west, inner sep=0]
        at (out3.south east) 
        {\tikz \drawModuleWithParams{$f^{(\ell)}$}{16}{$\theta^{(\ell)}$}{$\delta \theta^{(\ell)}$}{$\HeCal \theta^{(\ell)}$};};
    \node (out4)
        [inner sep=0, anchor=south west]
        at (layer5.south east)
        {\tikz \drawMessagesWithArrows{$z^{(\ell)}$}{$\delta  z^{(\ell)}$}{$\HeCal  z^{(\ell)}$}{\hNodeDistance};};

    \node (lossLayer) [inner sep=0pt, anchor=south west]
        at (out4.south east)
        {\tikz\drawModuleNoParams{$E$}{5};};
    \node (loss)
        [inner sep=0, anchor=south west]
        at (lossLayer.south east)
        {\tikz \drawMessagesWithArrows{$E$}{ }{ }{\hNodeDistance};};	
\end{tikzpicture}
 		}
		\caption{Extension of backprop to Hessians.
			It yields diagonal blocks 
			of the full parameter Hessian.
		}
		\label{fig:sketchFCNN}
	\end{figure*}
    
    Equation~\eqref{equ:hessianBackPropagation} is the central functional 
    expression herein, and will be referred to as the \emph{Hessian 
    backpropagation (HBP) equation}.
    Our suggested extension of gradient backpropagation is to also send
    the Hessian $\HeCal z$ back through the graph. To do so, existing 
    modules have to be extended by the HBP equation:
    \emph{Given the Hessian $\HeCal z$ of the loss with respect to all module 
    outputs, an extended module has to extract the Hessians $\HeCal \theta, 
    \HeCal x$ by means of Equation~\eqref{equ:hessianBackPropagation}, and 
    forward the Hessian with respect to its input $\HeCal x$ to the parent 
    module which proceeds likewise.}   
    In this way, backprop of gradients can be extended to compute 
    curvature information in modules. This corresponds to BDAs 
    of the Hessian that ignore second-order partial derivatives of 
    parameters in different modules. 
    Figure~\ref{fig:sketchFCNN} shows the data flow. The computations 
    required in Equation~\eqref{equ:hessianBackPropagation} depend only on 
    \emph{local quantities} 
    that are, mostly, already being computed during gradient
    backpropagation.\footnote{
        By Fa\`a di Bruno's formula~\citep{johnson2002FaaDiBruno}
        higher-order derivatives of 
        function compositions are expressed recursively in terms of the 
        composites' 
        lower-order derivatives. Recycling these quantities can give 
        significant speedup compared to repeatedly applying first-order 
        auto-differentiation, which represents one key aspect of our 
        work.
    }
    
    Before we proceed, we highlight the following aspects:
    \begin{itemize}
        \item The BDA of the Hessian need not be PSD. But our scheme can 
        be modified to provide PSD curvature matrices by projection onto the positive semi-definite cone (see Subsection~\ref{subsec:curvatureMatrices}). 
        
        \item 
        Instead of 
        evaluating all matrices during backpropagation, we can define 
        matrix-vector products recursively. This yields exact 
        curvature matrix products with the block diagonals of the Hessian, 
        the generalized Gauss-Newton (GGN) matrix and the PCH. Products with 
        the first two matrices can also be 
        obtained by use of automatic differentiation 
        \citep{pearlmutter1994FastExactHessianMultiplication, 
        schraudolph2002FastMatrixVectorProducts}. We also get access 
        to the PCH which, in contrast to 
        the GGN, 
        considers curvature information introduced by the 
        network (see 
        Subsection~\ref{subsec:curvatureMatrices}).\footnote{Implementations 
        of HBP for exact matrix-vector products 
        can reuse multiplication by the (transposed) Jacobian 
        provided by many machine learning libraries. 
        The second term of~\eqref{equ:hessianBackPropagation} needs 
        special treatment though.} 
        For standard neural networks, only second derivatives of nonlinear 
        activations have to be stored compared to gradient backpropagation.
        
        \item There are approaches \citep{botev2017PracticalGaussNewton, 
        chen2018BDAPCH} that propagate matrix representations 
        back through the graph in order to save repeated computations in the 
        curvature matrix-vector product. The 
        size of the matrices $\HeCal z^{(i)}$ passed between layer $i+1$ and 
        $i$ scales quadratically in 
        the number of output features of layer $i$. For convolutional layers 
        and in case of batched input data, the dimension of these quantities 
        exceeds computational budgets. In line with previous schemes 
        \citep{botev2017PracticalGaussNewton, chen2018BDAPCH}, we introduce 
        additional approximations for batch learning 
        in Subsection~\ref{subsec:batchLearning}. A connection to existing 
        schemes is drawn in the Supplements~\ref{subsec:relation}.   
    \end{itemize}
     
    HBP can easily be integrated into current machine 
    learning libraries, so that BDAs of curvature information can be provided 
    automatically for novel or existing second-order optimization methods. 
    Such methods have repeatedly been shown 
    to be competitive with first-order methods \citep{martens2015KFAC, 
    grosse2016KFACConvolution, botev2017PracticalGaussNewton, 
    zhang2017BlockDiagonalHessianFree, 
    chen2018BDAPCH}. 

    \paragraph{Relationship to matrix differential calculus:}
    \label{sec:MDF}
    To some extent, this paper is a re-formulation of 
    earlier results \citep{martens2015KFAC, 
    botev2017PracticalGaussNewton, chen2018BDAPCH} in the framework of matrix 
    differential calculus \citep{magnus1999MatrixDifferentialCalculus},
    leveraged to achieve a new level of modularity. Matrix 
    differential calculus is a set of notational rules that allow a concise 
    construction of derivatives without the heavy use of indices. 
    Equation~\eqref{equ:hessianBackPropagation} is a special 
    case of the matrix chain rule of that framework. A more detailed 
    discussion of this connection can be found in Section 
    \ref{sec:matrixDifferentialCalculus} of the Supplements, which also 
    reviews definitions generalizing the concepts of Jacobian and Hessian in 
    a way that preserves the chain rule. The 
    elementary building block of our procedure is a \emph{module} as shown in 
    Figure~\ref{fig:sketchModule}. Like for gradient backprop, the 
    operations required for HBP can be tabulated. Table 
    \ref{table:backpropEquations} provides a selection of common  
    modules. The derivations, which again 
    leverage the matrix differential calculus framework, can be found in 
    Supplements~\ref{sec:examples_fcnn}, \ref{sec:examples_loss}, and 
    \ref{sec:examples_cnn}.

    \begin{table*}[h]
        \caption{Hessian backpropagation for common modules used in feedforward networks.
            $I$ denotes the 
            identity 
            matrix. We assign matrices to upper-case ($W, X, \dots$) and 
            tensors to upper-case sans serif symbols ($\tensor{W}, 
            \tensor{X}, \dots$).}
        \label{table:backpropEquations}
        \centering
        \begin{tabular}{lll}
            \textbf{OPERATION} & \textbf{FORWARD} & \textbf{HBP} (Equation 
            \eqref{equ:hessianBackPropagation}) 
            \\
            \midrule
            Matrix-vector & $z(x, W) = Wx$ & $\HeCal x = W^\top (\HeCal 
            z) W$\,, 
            \\
            multiplication& & $\HeCal W = x \otimes x^\top \otimes \HeCal z$
            \\
            Matrix-matrix & $Z(X, W) = WX$ & $\HeCal X = (I \otimes 
            W)^\top \HeCal Z ( I \otimes W)$\,, 
            \\
            multiplication& & $\HeCal W = (X^\top \otimes I)^\top \HeCal Z 
            (X^\top \otimes 
            I)$
            \\	
            Addition & $z(x, b) = x + b$ & $\HeCal x = \HeCal b =\HeCal z $ 
            \\
            Elementwise & $z(x) = \phi(x)$\,, & $\HeCal x = 
            \diag[\phi'(x)]  \HeCal z \diag[\phi'(x)] + \diag[\phi''(x) 
            \odot \delta z]$
            \\
            activation& $z_i(x) = \phi(x_i)$ 
            \\
            \midrule
            Skip-connection & $z(x, \theta) = x + y(x, \theta)$ & $\HeCal x = 
            [I + \D y(x)]^\top \HeCal z [I + \D y(x)] 
            + \sum_k [\He y_k(x)] \delta z_k$\,, 
            \\
            & & $\HeCal \theta = [\D y(\theta)]^\top \HeCal z [\D y(\theta)]
            + \sum_k [\He y_k(\theta)] \delta z_k$ 
            \\	
            \midrule	
            Reshape/view & $\tensor{Z}(\tensor{X})= 
            \mathrm{reshape}(\tensor{X})$ & $\HeCal \tensor{Z} = \HeCal 
            \tensor{X}$ 
            \\
            Index select/map $\pi$ & $z(x) = \Pi x\, ,$ $\Pi_{j,\pi(j)} = 
            1\,, $ & $\HeCal x = \Pi^\top(\HeCal z)\Pi$ 
            \\
            Convolution & $\tensor{Z}(\tensor{X}, \tensor{W}) = \tensor{X} 
            \star \tensor{W}$\,, & $\HeCal \llbracket \tensor{X} \rrbracket = 
            (I \otimes W) \HeCal Z (I \otimes W)$ 
            \\ 
            & $Z(W, \llbracket\tensor{X}\rrbracket) = W 
            \llbracket \tensor{X} \rrbracket$\,, & $\HeCal W = (\llbracket 
            \tensor{X} \rrbracket^\top \otimes I)^\top \HeCal Z (\llbracket 
            \tensor{X} \rrbracket^\top \otimes I)$
            \\
            \midrule	
            Square loss & $E(x, y) = (y-x)^\top (y - x)$ & $\HeCal x = 2 I$ 
            \\
            Softmax cross-entropy & $E(x, y) = - y^\top \log\left[ 
            p(x)\right]$ & $\HeCal x = \diag\left[p(x)\right]- p(x) 
            p(x)^\top$ 
            \\
        \end{tabular}
    \end{table*}

\subsection{Obtaining different curvature matrices}
\label{subsec:curvatureMatrices}
	The HBP equation 
    yields 
	\emph{exact} diagonal 
	blocks $\HeCal \theta^{(1)} , \dots, \HeCal \theta^{(\ell)}$  of the 
	full parameter Hessian. 
    They can be of interest in their own right for analysis of the loss 
    function, but are not generally suitable for second-order optimization in 
    the sense of~\eqref{equ:NewtonUpdate}, as they need neither be PSD nor 
    invertible.
    For application in optimization, HBP can be modified to yield 
	semi-definite BDAs of the Hessian.
	Equation~\eqref{equ:hessianBackPropagation} again provides the foundation for 
	this adaptation, which is closely related to the concepts of 
	KFRA~\citep{botev2017PracticalGaussNewton}, 
	BDA-PCH~\citep{chen2018BDAPCH}, 
	and, under certain conditions, KFAC \citep{martens2015KFAC}. We draw 
	their connections by briefly reviewing them here.

	\paragraph{Generalized Gauss-Newton matrix:} The GGN emerges 
	as the curvature matrix in the quadratic expansion of the loss function 
	$E(z^{(\ell)})$ in terms of the network output $z^{(\ell)}$. 
	It is also obtained by linearizing the network output 
	$z^{(\ell)}(\theta, x)$ in $\theta$ before computing the loss 
	Hessian~\citep{martens2014NaturalGradient}, and reads
	\begin{align*}
		\mathrm{G}(\theta) = \frac{1}{|Q|} \sum_{(x, y) \in Q} \left[\D z^{(\ell)}(\theta)\right]^\top \He E(z^{(\ell)}) \left[\D z^{(\ell)}(\theta) \right]\,.
	\end{align*}
    To obtain diagonal blocks $\G(\theta^{(i)})$,
    the Jacobian
    can be unrolled by means of 
    the chain rule for Jacobians (Supplements, Theorem 
    \ref{the:chainRuleJacobians}) as
    \begin{math}
        \D z^{(\ell)}(\theta^{(i)}) = \left[\D z^{(\ell)}(z^{(\ell - 
        1)})\right] \left[\D z^{(\ell - 1)}(\theta^{(i)})\right] \dots
    \end{math}
    Continued expansion shows that the Hessian $\He E(z^{(\ell)})$ 
    of the loss function with respect to the network output is 
    propagated back through a layer by multiplication from left and 
    right with its Jacobian.
    This is accomplished in HBP by \emph{ignoring second-order effects 
    introduced by modules}, that is by setting the Hessian of 
    the module function to zero, therefore neglecting the second term in 
    Equation~\eqref{equ:hessianBackPropagation}. In fact, if all activations 
    in the network 
    are piecewise linear (e.g.~ReLUs),
    the GGN and Hessian blocks are
    equivalent. Moreover, diagonal blocks of the GGN 
    are PSD if the loss function is convex (and thus $\He E(z^{(\ell)})$ 
    is PSD). This is because blocks are recursively left- and 
    right-multiplied with Jacobians, which does not alter the definiteness. 
    Hessians of the loss functions listed in 
    Table~\ref{table:backpropEquations} are PSD.
    The resulting recursive scheme has been used by 
    \citet{botev2017PracticalGaussNewton} under the acronym KFRA to optimize 
    convex loss functions of fully-connected neural networks with piecewise 
    linear activation functions.
		
	 \paragraph{Positive-curvature Hessian:} Another concept of 
	 positive semi-definite BDAs of the Hessian (that additionally considers
         second-order module effects) was studied in \citet{chen2018BDAPCH} 
         and 
	 named the PCH. It is obtained by 
	 modification of terms in the second summand of 
	 Equation~\eqref{equ:hessianBackPropagation} that can 
	 introduce concavity 
	 during HBP. This ensures positive semi-definiteness since the first 
	 summand is semi-definite, assuming the loss Hessian $\He 
	 E(z^{(\ell)})$ with respect to the network output to be positive semi-definite. 
	 \citet{chen2018BDAPCH} suggest to eliminate negative 
	 curvature of a matrix by 
	 computing the eigenvalue decomposition and either discard negative 
	 eigenvalues or cast them to their absolute value.
     This allows the construction of PSD curvature matrices even for 
     non-convex loss functions. In the setting of~\citet{chen2018BDAPCH}, the 
     PCH can empirically outperform optimization using the GGN. In usual 
     feedforward neural networks, the concavity is introduced by nonlinear 
     elementwise activations, and corresponds to a diagonal matrix 
     (Table~\ref{table:backpropEquations}). Thus, convexity can be 
     maintained during HBP by either clipping negative values to zero 
     (PCH-clip), or 
     taking their magnitude in the diagonal concave term (PCH-abs).

	\paragraph{Fisher information matrix:} If the network defines a 
	conditional probability density $r(y | z^{(\ell)})$ on the labels, 
	maximum likelihood learning for the parameterized density $p_\theta (y | 
	x)$ will correspond to choosing a negative log-likelihood loss function, 
	i.e.~$E(z^{(\ell)}, y) = - \log r(y | z^{(\ell)})$.
    Common loss functions like square and cross-entropy loss 
    can be interpreted in this way. Natural gradient descent 
    \citep{amari1998NaturalGradient} uses the Fisher information matrix
	\begin{math}
		\F(\theta) = \E_{p_\theta(y | x)}\left[\left(\nicefrac{\diff \log 
		p_\theta(y | x)}{\diff \theta} \right) \left(\nicefrac{\diff \log 
		p_\theta(y | x)}{\diff \theta^\top} \right) \right]
    \end{math}
    as a PSD curvature matrix approximating the Hessian.
    It can be expressed as the log-predictive 
    density's expected Hessian 
    under $r$ itself: $F_r(z^{(\ell)}) = - 
    \E_{r(y | z^{(\ell)})}\left[ \He \log r(y | z^{(\ell)}) \right]$. 
    Assuming truly i.i.d.\ samples $x$, the log-likelihood of multiple 
    data decomposes and results in the approximation
    \begin{align*}
        \F (\theta) &\approx \dfrac{1}{|Q|} \sum_{(x, y) \in Q} \left[ \D 
        z^{(\ell)}(\theta)\right]^\top F_r(z^{(\ell)}) \left[\D 
        z^{(\ell)}(\theta)\right]\,.
    \end{align*}
    In this form, the computational scheme for BDAs of the Fisher 
    resembles the HBP of the GGN. However, instead of propagating back the 
    loss Hessian with respect to the network, the expected Hessian of the 
    negative log-likelihood under the model's predictive distribution is 
    used. 
    \citet{martens2015KFAC} use Monte-Carlo sampling to estimate this matrix 
    in their KFAC optimizer. 
    Relations between the Fisher and GGN are discussed in \citep{pascanu2013revisiting, martens2014NaturalGradient};
    for square and cross-entropy loss, they are equivalent.  
\subsection{Batch learning approximations}
\label{subsec:batchLearning}

    In our HBP framework, exact multiplication by the block of the 
    curvature matrix of parameter $\theta$ in a module comes at the 
    cost of one gradient backpropagation to this layer. The 
    multiplication is recursively defined in terms of multiplication by the 
    layer output Hessian $\HeCal z$. If it were 
    possible to have an explicit representation of this matrix in memory, the 
    recursive 
    computations hidden in $\HeCal z$ could be saved during the 
    solution of the linear system implied by 
    Equation~\eqref{equ:NewtonUpdate}.
    Unfortunately, the size of the backpropagated exact matrices scales 
    quadratically in both the batch size\footnote{
    If samples are processed independently in every module, these 
    matrices have block 
    structure and scale 
    linearly in batch size. Quadratic scaling is caused by transformations 
    across different samples, like batch normalization.    
} and the number of layer's output features. However, instead of 
    propagating back the exact Hessian, a batch-averaged version can be used 
    instead to circumvent the batch size scaling (originating 
    from \citet{botev2017PracticalGaussNewton}).
    In combination with structural information about the parameter 
    Hessian, this strategy is used in \citet{    
    botev2017PracticalGaussNewton, chen2018BDAPCH} to further approximate 
    curvature multiplications, using quantities computed 
    in a single backward pass and then kept in memory for application of 
    the matrix-vector product. We can embed these explicit schemes into 
    our modular approach.
    To do so, we denote averages over a batch $B$ by a bar, for instance 
    \begin{math}
        \nicefrac{1}{|B|} \sum_{(x, y) \in B} \He E(\theta) = \average{\He 
        E(\theta)}.
    \end{math}
    The modified backward pass of curvature information during HBP for a 
    module receives a batch average 
    of the Hessian with respect to the output, $\average{\HeCal z}$, which is 
    used to formulate the matrix-vector product with the batch-averaged 
    parameter Hessian $\average{\HeCal \theta}$. 
    An average of 
    the Hessian with respect to the module input, $\average{\HeCal x}$, is 
    passed back.
    Existing work \citep{botev2017PracticalGaussNewton, 
    chen2018BDAPCH} differs primarily in the specifics of how this batch 
    average is computed. In HBP, these 
    approximations can be formulated compactly within 
    Equation~\eqref{equ:hessianBackPropagation}.  
    Relations to the cited works are discussed in more detail in the 
    Supplements~\ref{subsec:relation}.
    The approximations amounting to relations used by 
    \citet{botev2017PracticalGaussNewton} read
    \begin{align}
        \label{equ:hessians_batch_average}
        \average{\HeCal x} \approx \average{\left[\D z(x)\right]^\top 
        \average{\HeCal z} \left[\D z(x)\right]}
        + \sum_k \average{\left[\He z_k(x)\right] \delta z_k}\,,
    \end{align}
    and likewise for $\theta$. In case of a linear layer $z(x) = Wx + b$, 
    this approximation implies the relations 
    $\average{\HeCal W} = \average{x \otimes x^\top} \otimes 
    \average{\HeCal z}$, $\average{\HeCal b} = \average{\HeCal z}$, and 
    $\average{\HeCal x} = W^\top 
    (\average{\HeCal z}) W$.
    Multiplication by this weight Hessian approximation with a vector $v$ 
    is achieved by storing $\average{ x \otimes x^\top}$, $\average{\HeCal 
    z}$ and performing the required contractions $v\mapsto 
    (\average{x \otimes x^\top} \otimes \average{\HeCal z}) v $. Note that 
    this approach is not restricted to curvature matrix-vector multiplication 
    routines only. 
    Kronecker structure in the approximation gives rise to 
    optimization methods relying on direct inversion.
    
     A cheaper approximation, used in~\citet{chen2018BDAPCH},
    \begin{align}
        \label{equ:hessians_batch_average_approximation}
        \average{\HeCal x} &\approx \average{\left[\D z(x)\right]}^\top 
        \average{\HeCal z}\ \, \average{\left[\D z(x)\right]} + \sum_k 
        \average{\left[\He z_k(x)\right] \delta z_k}\,,
    \end{align} 
    leads to the modified relation $\average{\HeCal W} = \average{x} 
    \otimes \average{x}^\top \otimes 
    \average{\HeCal z}$ for a linear layer. As this 
    approximation is of the same rank as $\average{\HeCal z}$, which is 
    typically small, CG requires only a few iterations during 
    optimization. It avoids large memory requirements for layers with numerous 
    inputs, since it requires $\average{x}$ be 
    stored instead of $\average{x\otimes x^\top}$.
    
    Transformations that are linear in the module parameters (e.g.\ linear 
    and convolutional layers), possess constant Jacobians with respect to the 
    module input for each sample (see Table~\ref{table:backpropEquations}). 
    Hence, 
    in a network consisting of only these layers, both 
    Equation~\eqref{equ:hessians_batch_average} 
    and~\eqref{equ:hessians_batch_average_approximation} yield the same 
    backpropagated Hessians $\average{\HeCal x}$. This still leaves the 
    degree of freedom for choosing the approximation scheme in the analogous 
    equations for $\theta$.
    
    \paragraph{Remark:}
    Both strategies for 
    obtaining curvature matrix BDAs (implicit exact 
    matrix-vector 
    multiplications and explicit 
    propagation of approximated curvature) are compatible. 
    Regarding the connection to cited works, we note that the 
    maximally modular structure of our framework 
    changes the nature of these approximations 
    and allows a more flexible formulation.

\section{Experiments \& Implementation}
\label{sec:experiments}

We illustrate the usefulness of incorporating curvature information with the 
two outlined strategies by 
experiments with a fully-connected and a convolutional neural network (CNN) 
on the 
CIFAR-10 dataset~\citep{krizhevsky2009CIFAR10}. 
Following the guidelines of 
\citet{schneider2018deepobs}, the
training loss is estimated on a random subset of the training set of equal 
size as the 
test set. Each experiment is performed for 10 different random seeds and we 
show the mean values with shaded intervals of one standard deviation. For the 
loss function we use cross-entropy. Details on the model architectures 
and hyperparameters are given in 
Supplements~\ref{sec:experimentalDetails}.

\paragraph{Training procedure and update rule:}

In comparison to a first-order optimization procedure, the training loop with 
HBP has to be extended by a single backward pass to backpropagate the 
batch-averaged or exact loss Hessian. This yields matrix-vector products with 
a curvature estimate 
$C^{(i)}$ for each parameter block $\theta^{(i)}$ of the network. 
Parameter updates $\Delta \theta^{(i)}$ are obtained by applying CG to solve 
the linear system\footnote{We use the same update rule as 
\citet{chen2018BDAPCH} since we extend some of the results shown within this 
work.}
\begin{align}
    \label{equ:linearSystemCG}
    \left[ \alpha I + (1 - \alpha) C^{(i)} \right]  \Delta 
    \theta^{(i)} = - \delta \theta^{(i)}\,,
\end{align} 
where $\alpha$ acts as a step size limitation to improve robustness 
against noise. The CG routine terminates if the ratio of 
the residual norm and the gradient norm falls below a certain threshold or 
the maximum number of iterations has been reached. The solution returned by 
CG is scaled by a learning rate $\gamma$, and parameters are updated by the 
relation $\theta^{(i)} \leftarrow \theta^{(i)} + \gamma \Delta \theta^{(i)}.$

\begin{figure}[!t]
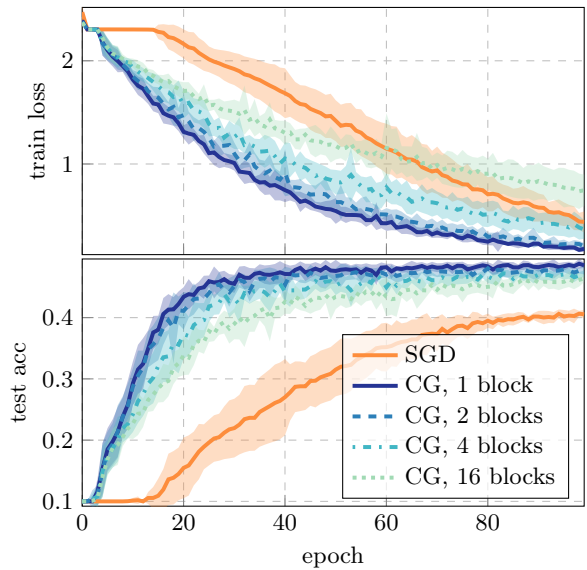

	\centering
	\footnotesize
	\setlength{\figwidth}{\linewidth}
	\setlength{\figheight}{0.59\figwidth}
	\resetPGFStyle
	\pgfkeys{/pgfplots/mystyle/.style={
			original,
			legend pos = north east,
			ylabel near ticks,
			xlabel near ticks,
			xlabel style={opacity=0},
			xticklabel style={opacity=0, yshift=3ex},
			legend style = {fill opacity = 0, 
				text opacity = 0,
				draw opacity = 0},   
	}}
   	\begin{center}
     \end{center}	
	\vspace{-6.3ex}
	\resetPGFStyle
	\pgfkeys{/pgfplots/mystyle/.style={
			original,
			legend pos = south east, 
			ylabel near ticks,
			xlabel near ticks,  
	}}
	\begin{center}
		\hspace*{-0.23cm}%
 	\end{center}	
	\vspace{-2ex}
	\caption{SGD and 
		different Newton-style 
		optimizers based on the PCH-abs
		with batch approximations. The same fully-connected neural network 
		of~\citep{chen2018BDAPCH} was used to generate the solid baseline 
		results. Our modular approach allows further splitting the parameter 
		blocks into sub-blocks that can independently be optimized in 
		parallel (dashed lines).}
	\label{fig:experiment_fcnn}  
\end{figure}

\paragraph{Fully-connected network, batch approximations, and sub-blocking:}
The flexibility of HBP is illustrated by extending the results in 
\citet{chen2018BDAPCH}. Investigations are 
performed on a fully-connected network with sigmoid activations. Solid 
lines in Figure \ref{fig:experiment_fcnn} show the performance of the 
Newton-style optimizer and momentum SGD in terms of the training loss and 
test accuracy. The 
second-order method is capable to escape the initial plateau in 
fewer iterations.

\begin{figure*}[!t]
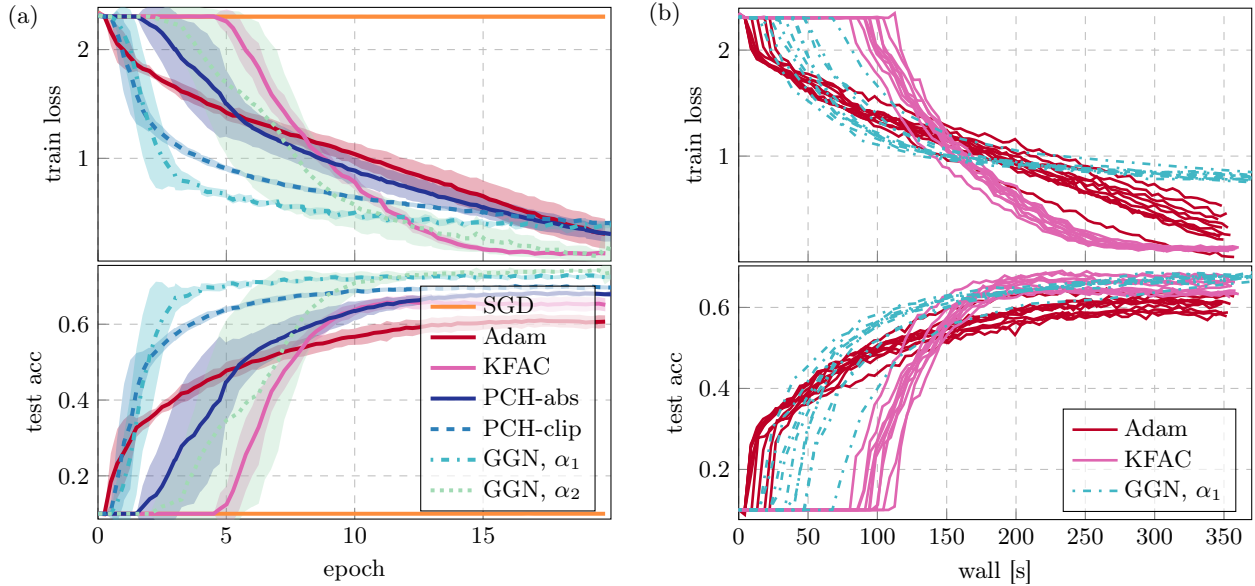

	\centering	
	\begin{minipage}{0.49\linewidth}
		\footnotesize
		\setlength{\figwidth}{\linewidth}
		\setlength{\figheight}{0.59\figwidth}
		\resetPGFStyle
		\pgfkeys{/pgfplots/mystyle/.style={
				original,
				legend pos = north east,
				ylabel near ticks,
				xlabel near ticks,
				xlabel style={opacity=0},
				xticklabel style={opacity=0, yshift=3ex},  
				every axis plot/.append style={ultra thick},
				legend style = {fill opacity = 0, 
					text opacity = 0,
					draw opacity = 0},   
		}}
		\begin{flushleft}
			(a)
		\end{flushleft}
		\vspace{-5ex}
		\begin{center}
 		\end{center}
		\vspace{-6.3ex}
		\resetPGFStyle
		\pgfkeys{/pgfplots/mystyle/.style={
				original,
				legend pos = south east, 
				ylabel near ticks,
				xlabel near ticks,
		}}
		\begin{center}
			\hspace*{-0.23cm}%
 		\end{center}
	\end{minipage}
	\begin{minipage}{0.49\linewidth}
		\footnotesize
		\begin{flushleft}
			(b)
		\end{flushleft}
		\vspace{-4.6ex}
		\setlength{\figwidth}{\linewidth}
		\setlength{\figheight}{0.59\figwidth}
		\resetPGFStyle
		\pgfkeys{/pgfplots/mystyle/.style={
				original,
				xmin = 0, xmax = 370,
				every axis plot/.append style={line width = 1pt},
				legend style = {fill opacity = 0, 
					text opacity = 0,
					draw opacity = 0}, 
				ylabel near ticks,
				xlabel near ticks,
				xlabel style={opacity=0},
				xticklabel style={opacity=0, yshift=3ex},    
		}}
		\begin{center}
			%
 		\end{center}
		\vspace{-6.6ex}	
		\resetPGFStyle
		\pgfkeys{/pgfplots/mystyle/.style={
				original,
				xmin = 0, xmax = 370,
				ylabel near ticks,
				xlabel near ticks,
				every axis plot/.append style={line width = 1pt},
				legend pos = south east,
		}}
		\begin{center}
			\hspace*{-0.23cm}%
 		\end{center}
	\end{minipage}
	\resetPGFStyle
	\vspace{-2ex}
	\caption{(a) Comparison of SGD, Adam, KFAC, and Newton-style 
		methods with different exact curvature matrix-vector products (HBP) 
		on a CNN with sigmoid activations
		(see Supplements~\ref{sec:experimentalDetails}). SGD cannot train the net. (b) Wall-clock time
		comparison (on an RTX 2080 Ti GPU; same colors realize 
		different random seeds).}
	\label{fig:experiment_cnn}        
\end{figure*}

The modularity of HBP allows for additional parallelism by splitting the 
linear system \eqref{equ:linearSystemCG} into smaller sub-blocks, which then 
also need fewer iterations of CG. Doing so only requires a minor modification 
of the parameter Hessian computation by~\eqref{equ:hessianBackPropagation}. 
Consequently, 
we split weights and bias terms row-wise into a specified number of 
sub-blocks. Performance curves are shown in 
Figure~\ref{fig:experiment_fcnn}.
In the initial phase, the BDA can be split into 
a larger number of sub-blocks without suffering from a loss in performance. 
The reduced curvature information is still sufficient to escape the initial 
plateau. However, larger blocks have to be considered in later 
stages to further reduce the loss efficiently. 

The fact that this switch in modularity is necessary is an argument 
in favor of the flexible form of HBP, which allows to efficiently realize 
such switches: For this experiment, the splitting for each block 
was artificially chosen to illustrate this flexibility. In 
principle, the splitting could be decided individually for each parameter 
block, and even changed at run time.

\paragraph{Convolutional neural network, matrix-free exact curvature multiplication:} 
For convolutions, the large number of hidden 
features prohibits backpropagating a curvature matrix batch average. 
Instead, we use exact curvature matrix-vector products 
provided within HBP. The CNN possesses sigmoid activations and 
cannot be trained by SGD (cf.~Figure~\ref{fig:experiment_cnn}a). For 
comparison with another second-order method, we experiment with 
a public KFAC implementation~\citep[see 
Supplements~\ref{sec:experimentalDetails} for 
details]{martens2015KFAC, 
grosse2016KFACConvolution}. 

The matrix-free
second-order methods progress fast 
in the initial stage of the optimization. However, progress in later 
phases stagnates. This may be caused by the limited sophistication of the 
update 
rule~\eqref{equ:linearSystemCG}: If a small value for 
$\alpha$ is chosen, the optimizer will perform well in the beginning (GGN, 
$\alpha_1$). As the 
gradients become smaller, and hence more noisy, the step size limitation is 
too optimistic, which leads to a slow-down in optimization progress. A more 
conservative step size limitation improves the overall performance at the 
cost of fewer initial progress (GGN, $\alpha_2$). 
In the training 
phase where damping is ``effective'', our 
illustrative methods, and KFAC, exhibit better progress per 
iteration on the objective than the first-order competitor Adam, underlining 
the usefulness of 
curvature even if only computed block-wise. 

For an impression on performance in terms of run time, 
Figure~\ref{fig:experiment_cnn}b compares the wall-clock time of
one matrix-free method and the baselines. 
The HBP-based optimizer can compete with 
existing methods and offers potential for further improvements, like 
sub-blocking and parallelized CG.
Despite the more adaptive nature of second-order methods, their full power
seems to still require adaptive damping, to account for the quality 
of the local quadratic approximation and  restrict the 
update if necessary. The importance of these techniques to properly adapt the 
Newton direction has been emphasized in previous works 
\citep{martens2010HessianFree,martens2015KFAC, botev2017PracticalGaussNewton} that aim to 
develop fully fletched second-order optimizers.
Such adaptation, however, is beyond the scope of this 
text.
 
\section{Conclusion}
We have outlined a procedure to compute block-diagonal approximations of 
different curvature matrices for feedforward neural networks by a scheme that 
can be realized on top of gradient backpropagation. In contrast to other 
recently proposed methods, our implementation is 
aligned with the design of current machine learning frameworks and can 
flexibly compute Hessian sub-blocks to different levels of refinement. 
Its modular formulation facilitates closed-form analysis of 
Hessian diagonal blocks, and unifies previous approaches 
\citep{botev2017PracticalGaussNewton, chen2018BDAPCH}.

Within our framework we presented two strategies: (i) Obtaining exact curvature 
matrix-vector products that have not been 
accessible before by auto-differentiation (PCH), 
and (ii) backpropagation of further approximated matrix representations 
to save computations during training. As for gradient backpropagation, the Hessian backpropagation for 
different operations can be derived independently of the underlying graph. 
The extended modules 
can then be used as a drop-in replacement for existing modules to construct 
deep neural networks. 
Internally, backprop is extended by an additional Hessian 
backward pass through the graph to compute curvature information. It can 
be performed in parallel to, and reuse the
quantities computed in, gradient backpropagation.

\vfill 
\section*{Acknowledgments}
The authors would like to thank Frederik Kunstner, Matthias Werner, Frank Schneider, and Agustinus Kristiadi for their constructive feedback on the manuscript, and 
gratefully acknowledge financial support by the European Research Council through ERC StG Action 757275 / PANAMA; the DFG Cluster of Excellence “Machine Learning - New Perspectives for Science”, EXC 2064/1, project number 390727645; the German Federal Ministry of Education and Research (BMBF) through the T\"ubingen AI Center (FKZ: 01IS18039A); and funds from the Ministry of Science, Research and Arts of the State of Baden-W\"urttemberg. F.\,D.\,is grateful to the International Max Planck Research School for Intelligent Systems (IMPRS-IS) for support.

 \clearpage 

\section*{Supplementary material}

# Modular Block-diagonal Curvature Approximations for Feedforward Architectures
## (Supplementary Material)

**Felix Dangel**
University of Tübingen
fdangel@tue.mpg.de

**Stefan Harmeling**
Heinrich Heine University Düsseldorf
harmeling@hhu.de

**Philipp Hennig**
University of Tübingen and
MPI for Intelligent Systems, Tübingen
ph@tue.mpg.de

This document provides additional information and derivations to clarify the relations of Hessian backpropagation (HBP). Section A contains a clean definition of Jacobian and Hessian matrices for multi-variate functions that are typically involved in the construction of neural networks. We introduce conventions for notation and show how the HBP Equation (7) results naturally from the chain rule for matrix derivatives (Magnus and Neudecker, 1999).

With matrix derivatives, the HBP equation for a variety of module functions can be derived elegantly. Sections B, C, and D contain the HBP derivations for all operations in Table 1. We split the considered operations into different categories to achieve a cleaner structure. Section B contains details on operations used for the construction of fully-connected neural networks (FCNNs) and skip-connections. Subsection B.4 illustrates the analytic composition of multiple modules by combining the backward passes of a nonlinear elementwise activation function and an affine transformation. This yields the recursive schemes of Botev et al. (2017) and Chen et al. (2018), the latter of which has been used in the experiment of Section 4. The analysis of the Hessian for common loss functions is provided in Section C. Operations occurring in convolutional neural networks (CNNs) are subject of Section D.

Section E provides details on model architectures, training procedures used in the experiments of Section 4, and an additional experiment on a modified test problem of the DeepOBS benchmark library (Schneider et al., 2019).

## A   Matrix derivatives

Index notation for higher-order derivatives of multi-variate matrix functions can become heavy (Mizutani and Dreyfus, 2008; Naumov, 2017; Chen et al., 2018; Bakker et al., 2018). We tackle this issue by embedding the presented approach in the notation of matrix differential calculus, which

1. yields notation consistent with established literature on matrix derivatives (Magnus and Neudecker, 1999) and clarifies the origin of the symbols D and H that are used extensively in the main text.

2. allows for using a multi-dimensional generalization of the chain rule.

3. lets us extract first- and second-order derivatives from differentials using the identification rules of Magnus and Neudecker (1999) without bothering to deal with index notation.

With these techniques it is easy to see how structure, like Kronecker products, appears in the derivatives.

**Preliminaries & notation:**  The following definitions and theorems represent a collection of results from the book of Magnus and Neudecker (1999). They generalize the concept of first- and second-order derivatives to multi-variate matrix functions in terms of the Jacobian and Hessian matrix. While there exist multiple ways to arrange the partial derivatives, the presented definitions allows for a multivariate generalization of the chain rule.

We denote matrix, vector, and scalar functions by $F$, $f$, and $\phi$, respectively. Matrix (vector) arguments are written as $X$ ($x$). Vectorization (vec) applies column-stacking, such that for matrices $A, B, C$ of appropriate size

$$\vec(ABC) = \left(C^\top \otimes A\right) \vec(B). \tag{S.1}$$

Whenever possible, we assign vectors to lower-case (for instance $x, \theta$), matrices to upper-case ($W, X, \dots$), and tensors to upper-case sans serif symbols ($\mathsf{W}, \mathsf{X}, \dots$). $\odot$ denotes elementwise multiplication (Hadamard product).

**Remark on vectorization:** The generalization of Jacobians and Hessians provided by Magnus and Neudecker (1999) relies on vectorization of matrices. Convolutional neural networks usually act on tensors and we incorporate these by assuming them to be flattened such that the first index varies fastest. For a matrix (tensor of order two), this is consistent with column-stacking. For instance, the vectorized version of the tensor $\mathsf{A} \in \mathbb{R}^{n_1 \times n_2 \times n_3}$ with $n_1, n_2, n_3 \in \mathbb{N}$ is $\operatorname{vec} \mathsf{A} = (\mathsf{A}_{1,1,1}, \mathsf{A}_{2,1,1}, \dots, \mathsf{A}_{n_1,1,1}, \mathsf{A}_{1,2,1}, \dots, \mathsf{A}_{n_1,n_2,n_3})^\top$. To formulate the generalized Jacobian or Hessian for tensor operations, its action on a vector or matrix view of the original tensor is considered. Consequently, all operations can be reduced to vector-valued functions, which we consider in the following.

The vectorization scheme is not unique. Most of the linear algebra literature assumes column-stacking. However, when it comes to implementations, a lot of programming languages store tensors in row-major order, corresponding to row-stacking vectorization (last index varies fastest). Thus, special attention has to be paid in implementations.

### A.1 Definitions

**Definition A.1** (Jacobian matrix). Let $F: \mathbb{R}^{n \times q} \to \mathbb{R}^{m \times p}, X \mapsto F(X)$ be a differentiable function mapping between two matrix-sized quantitites. The *Jacobian* $\mathrm{D}F(X)$ of $F$ with respect to $X$ is an $(mp \times nq)$ matrix

$$\mathrm{D}F(X) = \frac{\partial \operatorname{vec} F(X)}{\partial (\operatorname{vec} X)^\top}, \qquad \text{such that} \qquad [\mathrm{D}F(X)]_{i,j} = \frac{\partial \left[\operatorname{vec} F(X)\right]_i}{\partial \left[(\operatorname{vec} X)^\top\right]_j} \tag{S.2}$$

(Magnus and Neudecker, 1999, Chapter 9.4).

In the context of FCNNs, the most common occurrences of Definition A.1 involve vector-to-vector functions $f: \mathbb{R}^n \to \mathbb{R}^m, x \mapsto f(x)$ with

$$\mathrm{D}f(x) = \frac{\partial f(x)}{\partial x^\top}.$$

For instance, $x$ can be considered the input or bias vector of a layer applying an affine transformation. Other cases involve matrix-to-vector mappings $f: \mathbb{R}^{n \times q} \to \mathbb{R}^m, X \mapsto f(X)$ with

$$\mathrm{D}f(X) = \frac{\partial f(X)}{\partial (\operatorname{vec} X)^\top},$$

where $X$ might correspond to the $\mathbb{R}^{m \times q}$ weight matrix of an affine transformation.

Proper arrangement of the partial derivatives leads to a generalized formulation of the chain rule.

**Theorem A.1** (Chain rule for Jacobians). Let $F: \mathbb{R}^{n \times q} \to \mathbb{R}^{m \times p}$ and $G: \mathbb{R}^{m \times p} \to \mathbb{R}^{r \times s}$ be differentiable matrix-to-matrix mappings and their composition $H$ be given by $H = G \circ F: \mathbb{R}^{n \times q} \to \mathbb{R}^{r \times s}$, then

$$\mathrm{D}H(X) = [\mathrm{D}G(F)] \, \mathrm{D}F(X) \tag{S.3}$$

(restricted from Magnus and Neudecker, 1999, Chapter 5.15).

Theorem A.1 is used to unroll the Jacobians in the composite structure of the feedforward neural network's loss function $E \circ f^{(\ell)} \circ f^{(\ell-1)} \circ \dots f^{(1)}$, compare Subsection 3.1.

**Definition A.2** (Hessian). Let $F: \mathbb{R}^{n \times q} \to \mathbb{R}^{m \times p}$ be a twice differentiable matrix function. The *Hessian* $\mathrm{H}F(X)$ is an $(mnpq \times nq)$ matrix defined by

$$\mathrm{H}F(X) = \mathrm{D}\left[\mathrm{D}F(X)\right]^\top = \frac{\partial}{\partial (\operatorname{vec} X)^\top} \operatorname{vec} \left\{ \left[\frac{\partial \operatorname{vec} F(X)}{\partial (\operatorname{vec} X)^\top}\right]^\top \right\} \tag{S.4}$$

(Magnus and Neudecker, 1999, Chapter 10.2).

Forms of Equation (S.4) that are most likely to emerge in neural networks include

$$\mathrm{H}\phi(x) = \frac{\partial^2 \phi(x)}{\partial x^\top \partial x} \quad \text{and} \quad \mathrm{H}\phi(X) = \frac{\partial}{\partial (\operatorname{vec} X)^\top} \frac{\partial \phi(X)}{\partial \operatorname{vec} X}.$$

The scalar function $\phi$ can be considered as the loss function $E$. Analogously, the Hessian matrix of a vector-in-vector-out function $f$ reads

$$\mathrm{H}f(x) = \begin{pmatrix} \mathrm{H}f_1(x) \\ \vdots \\ \mathrm{H}f_m(x) \end{pmatrix}. \tag{S.5}$$

Arranging the partial second derivatives in the fashion of Definition A.2 yields a direct generalization of the chain rule for second derivatives. We now provide this rule for a composition of vector-to-vector functions.

**Theorem A.2** (Chain rule for Hessian matrices). Let $f: \mathbb{R}^n \to \mathbb{R}^m$ and $g: \mathbb{R}^m \to \mathbb{R}^p$ be twice differentiable and $h = g \circ f: \mathbb{R}^n \to \mathbb{R}^p$. The relation between the Hessian of $h$ and the Jacobians and Hessians of the constituents $f$ and $g$ is given by

$$\mathrm{H}h(x) = [I_p \otimes \mathrm{D}f(x)]^\top [\mathrm{H}g(f)] \mathrm{D}f(x) + [\mathrm{D}g(f) \otimes I_n] \mathrm{H}f(x) \tag{S.6}$$

(restricted from Magnus and Neudecker, 1999, Chapter 6.10).

### A.2 Relation to the modular approach

Theorem A.2 can directly be applied to the graph $E \circ f^{(\ell)} \circ f^{(\ell-1)} \circ \cdots \circ f^{(1)}$ of the feedforward net under investigation. Note that, for any module function $f^{(i)}$, the loss can be expressed as a composition of two functions by squashing preceding modules in the graph into a single function $f^{(i-1)} \circ \cdots \circ f^{(1)}$, and likewise composing the module itself and all subsequent functions, i.e. $E \circ f^{(\ell)} \circ \cdots \circ f^{(i)}$.

The analysis can therefore be reduced to the module shown in Figure 2 receiving an input $x \in \mathbb{R}^n$ that is used to compute the output $z \in \mathbb{R}^m$. The scalar loss is then expressed as a mapping $E(z(x), y): \mathbb{R}^n \to \mathbb{R}^p$ with $p = 1$. Suppressing the label $y$, Equation (S.6) implies

$$\begin{aligned} \mathrm{H}E(x) &= [I_p \otimes \mathrm{D}z(x)]^\top [\mathrm{H}E(z)] \mathrm{D}z(x) + [\mathrm{D}E(z) \otimes I_n] \mathrm{H}z(x) \\ &= [\mathrm{D}z(x)]^\top [\mathrm{H}E(z)] \mathrm{D}z(x) + [\mathrm{D}E(z) \otimes I_n] \mathrm{H}z(x). \end{aligned} \tag{S.7}$$

The HBP Equation (7) is obtained by substituting (S.5) into (S.7).

## B HBP for fully-connected neural networks

### B.1 Linear layer (matrix-vector multiplication, matrix-matrix multiplication, addition)

Consider the function $f$ of a module applying an affine transformation to a vector. Apart from the input $x$, additional parameters of the module are given by the weight matrix $W$ and the bias term $b$,

$$\begin{aligned} f: \quad & \mathbb{R}^n \times \mathbb{R}^{m \times n} \times \mathbb{R}^m \to \mathbb{R}^m \\ & (x, W, b) \mapsto z = Wx + b. \end{aligned}$$

To compute the Jacobians with respect to each variable, we use the differentials

$$\begin{aligned} \mathrm{d}z(x) &= W \mathrm{d}x, \\ \mathrm{d}z(b) &= \mathrm{d}b, \\ \mathrm{d}z(W) &= (\mathrm{d}W)x \implies \mathrm{d}\operatorname{vec} z(W) = \left(x^\top \otimes I_m\right) \operatorname{vec}(\mathrm{d}W), \end{aligned}$$

using Property (S.1) to establish the implication in the last line. With the *first identification tables* provided in Magnus and Neudecker (1999, Chapter 9.6), the Jacobians can be read off from the differentials as $\mathrm{D}z(x) =$

$W$, $\mathrm{D}z(b) = I_m$, $\mathrm{D}z(W) = x^\top \otimes I_m$. All second module derivatives $\mathrm{H}z(x)$, $\mathrm{H}z(W)$, and $\mathrm{H}z(b)$ vanish since $f$ is linear in all inputs. Inserting the Jacobians into Equation (7) results in

$$\mathcal{H}x = W^\top \mathcal{H}z W, \tag{S.8a}$$
$$\mathcal{H}b = \mathcal{H}z, \tag{S.8b}$$
$$\mathcal{H}W = \left(x^\top \otimes I_m\right)^\top \mathcal{H}z \left(x^\top \otimes I_m\right) = xx^\top \otimes \mathcal{H}z = x \otimes x^\top \otimes \mathcal{H}z. \tag{S.8c}$$

The HBP relations for matrix-vector multiplication and addition listed in Table 1 are special cases of Equation (S.8). HBP for matrix-matrix multiplication is derived in a completely analogous fashion.

### B.2 Elementwise activation

Next, consider the elementwise application of a nonlinear function,

$$\phi: \quad \mathbb{R}^m \to \mathbb{R}^m$$
$$x \mapsto z = \phi(x) \quad \text{such that} \quad \phi_k(x) = \phi(x_k),$$

For the matrix differential with respect to $x$, this implies

$$\mathrm{d}\phi(x) = \phi'(x) \odot \mathrm{d}x = \mathrm{diag}\left[\phi'(x)\right] \mathrm{d}x,$$

and consequently, the Jacobian is given by $\mathrm{D}\phi(x) = \mathrm{diag}\left[\phi'(x)\right]$. For the Hessian, note that the function value $\phi_k(x)$ only depends on $x_k$ and thus $\mathrm{H}\phi_k(x) = \phi''(x_k) e_k e_k^\top$, with the one-hot unit vector $e_k \in \mathbb{R}^m$ in coordinate direction $k$. Inserting all quantities into the Relation (7) results in

$$\begin{aligned}
\mathcal{H}x &= \mathrm{diag}\left[\phi'(x)\right] \mathcal{H}z \, \mathrm{diag}\left[\phi'(x)\right] + \sum_k \phi''(x_k) e_k e_k^\top \delta z_k \\
&= \mathrm{diag}\left[\phi'(x)\right] \mathcal{H}z \, \mathrm{diag}\left[\phi'(x)\right] + \mathrm{diag}\left[\phi''(x) \odot \delta z\right].
\end{aligned} \tag{S.9}$$

### B.3 Skip-connection

Residual learning He et al. (2016) uses skip-connection units to facilitate the training of deep neural networks. In its simplest form, the mapping $f: \mathbb{R}^m \to \mathbb{R}^m$ reads

$$z(x, \theta) = x + y(x, \theta),$$

with a potentially nonlinear operation $(x, \theta) \mapsto y$. The Jacobians with respect to the input and parameters are given by $\mathrm{D}z(x) = I_m + \mathrm{D}y(x)$ and $\mathrm{D}z(\theta) = \mathrm{D}y(\theta)$. Using (7), one finds

$$\mathcal{H}x = \left[I_m + \mathrm{D}y(x)\right]^\top \mathcal{H}z \left[I_m + \mathrm{D}y(x)\right] + \sum_k \left[\mathrm{H}y_k(x)\right] \delta z_k,$$
$$\mathcal{H}\theta = \left[\mathrm{D}y(\theta)\right]^\top \mathcal{H}z \left[\mathrm{D}y(\theta)\right] + \sum_k \left[\mathrm{H}y_k(\theta)\right] \delta z_k.$$

### B.4 Relation to recursive schemes in previous work

The modular decomposition of curvature backpropagation facilitates the analysis of modules composed of multiple operations. Now, we analyze the composition of two modules. This yields the recursive schemes presented by Botev et al. (2017) and Chen et al. (2018), referred to as KFRA and BDA-PCH, respectively.

#### B.4.1 Analytic composition of multiple modules

Consider the module $g = f \circ \phi, x \mapsto y = \phi(x), y \mapsto z = f(y(x))$. We assume $\phi$ to act elementwise on the input, followed by a linear layer $f: z(y) = Wy + b$ (Figure S.1a). Analytic elimination of the intermediate backward pass yields a single module composed of two operations (Figure S.1b).

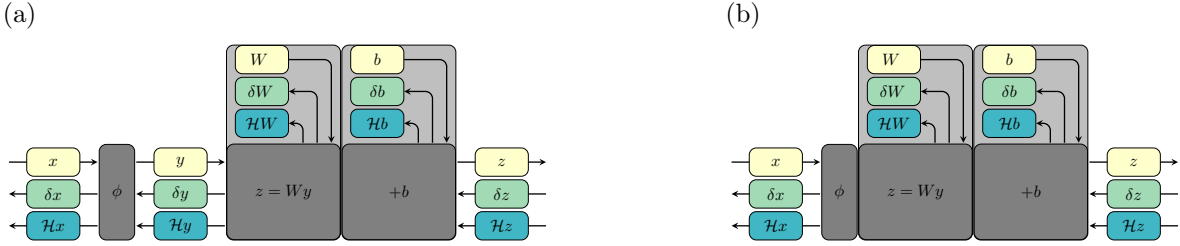

Figure S.1: Composition of elementwise activation $\phi$ and linear layer. (a) Both operations are analyzed separately to derive the HBP. (b) Backpropagation of $\mathcal{H}z$ is expressed in terms of $\mathcal{H}x$ without intermediate message.

The first Hessian backward pass through the linear module $f$ (Equations (S.8)) implies

$$\mathcal{H}y = W^\top \mathcal{H}z W,$$
$$\mathcal{H}W = y \otimes y^\top \otimes \mathcal{H}z = \phi(x) \otimes \phi(x)^\top \otimes \mathcal{H}z, \tag{S.10a}$$
$$\mathcal{H}b = \mathcal{H}z. \tag{S.10b}$$

Further backpropagation through $\phi$ by means of Equation (S.9) results in

$$\begin{aligned}
\mathcal{H}x &= \mathrm{diag}\left[\phi'(x)\right] \mathcal{H}y \,\mathrm{diag}\left[\phi'(x)\right] + \mathrm{diag}\left[\phi''(x) \odot \delta y\right] \\
&= \mathrm{diag}\left[\phi'(x)\right] \left[W^\top \mathcal{H}z W\right] \mathrm{diag}\left[\phi'(x)\right] + \mathrm{diag}\left[\phi''(x) \odot W^\top \delta z\right] \\
&= \{W \,\mathrm{diag}\left[\phi'(x)\right]\}^\top \mathcal{H}z \,\{W \,\mathrm{diag}\left[\phi'(x)\right]\} + \mathrm{diag}\left[\phi''(x) \odot W^\top \delta z\right].
\end{aligned} \tag{S.10c}$$

We use the invariance of a diagonal matrix under transposition and $\delta y = W^\top \delta z$ for the backpropagated gradient to establish the last equality. The Jacobian $\mathrm{D}g(x)$ of the module shown in Figure S.1b is $\mathrm{D}g(x) = W \,\mathrm{diag}\left[\phi(x)\right] = \left[W^\top \odot \phi'(x)\right]^\top$. In summary, the HBP relation for the composite layer $z(x) = W\phi(x) + b$ is given by (S.10).

### B.4.2  Obtaining the relations of KFRA and BDA-PCH

The derivations for the composite module given above are closely related to the recursive schemes of Botev et al. (2017); Chen et al. (2018). Their relations are obtained from a straightforward conversion of the HBP rules (S.10). Consider a sequence of a linear layer $f^{(1)}$ and multiple composite modules $f^{(2)}, \ldots, f^{(\ell)}$ as shown in Figure S.2.

According to Equation (S.10b) both the linear layer and the composite $f^{(i)}$ identify the gradient (Hessian) with respect to their outputs, $\delta z^{(i)}$ ($\mathcal{H}z^{(i)}$), as the gradient (Hessian) with respect to their bias term, $\delta b^{(i)}$ ($\mathcal{H}b^{(i)}$). Introducing layer indices for all quantities, one finds the recursion

$$\mathcal{H}b^{(i)} = \mathcal{H}z^{(i)}, \tag{S.11a}$$
$$\mathcal{H}W^{(i)} = \phi(z^{(i-1)}) \otimes \phi(z^{(i-1)})^\top \otimes \mathcal{H}b^{(i)}, \tag{S.11b}$$

for $i = \ell - 1, \ldots, 1$, and

$$\begin{aligned}
\mathcal{H}z^{(i-1)} &= \{W^{(i)} \,\mathrm{diag}\left[\phi'(z^{(i-1)})\right]\}^\top \mathcal{H}b^{(i)} \{W^{(i)} \,\mathrm{diag}\left[\phi'(z^{(i-1)})\right]\} + \mathrm{diag}\left[\phi''(z^{(i-1)}) \odot W^{(i)\top} \delta b^{(i)}\right] \\
&= \{W^{(i)\top} \odot \phi'(z^{(i-1)})\} \mathcal{H}b^{(i)} \{W^{(i)\top} \odot \phi'(z^{(i-1)})\}^\top + \mathrm{diag}\left[\phi''(z^{(i-1)}) \odot W^{(i)\top} \delta b^{(i)}\right]
\end{aligned} \tag{S.11c}$$

for $i = \ell - 1, \ldots, 2$. It is initialized with the gradient $\delta z^{(\ell)}$ and Hessian $\mathcal{H}z^{(\ell)}$ of the loss function.

Equations (S.11) are equivalent to the expressions provided in Botev et al. (2017); Chen et al. (2018). Their emergence from compositions of HBP equations of simple operations represents one key insight of this paper. Both works use the batch average strategy presented in Subsection 3.2 to obtain curvature estimates.

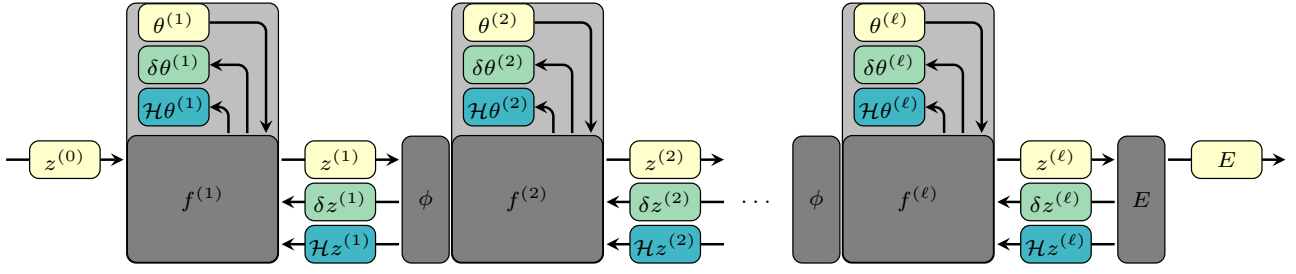

Figure S.2: Grouping scheme for the recursive Hessian computation proposed by KFRA and BDA-PCH. The backward messages between the linear layer and the preceding nonlinear activation are analytically absorbed.

## C  HBP for loss functions

### C.1  Square loss

Square loss of the model prediction $x \in \mathbb{R}^m$ and the true label $y \in \mathbb{R}^m$ is computed by

$$E(x, y) = (y - x)^\top (y - x).$$

Differentiating $\mathrm{d}E(x) = -(\mathrm{d}x)^\top (y - x) - (y - x)^\top \mathrm{d}x = -2(y - x)^\top \mathrm{d}x$ once more yields

$$\mathrm{d}^2 E(x) = 2(\mathrm{d}x)^\top \mathrm{d}x = 2(\mathrm{d}x)^\top I_m \mathrm{d}x.$$

The Hessian is extracted with the *second identification tables* from Magnus and Neudecker (1999, Chapter 10.4) and reproduces the expected result

$$\mathrm{H}E(x) = \mathcal{H}x = 2I_m. \tag{S.12}$$

### C.2  Softmax cross-entropy loss

The computation of cross-entropy from logits is composed of two operations. First, the neural network outputs are transformed into log-probabilities by the softmax function. Then, the cross-entropy with the label is computed.

**Log-softmax:** The output's elements $x \in \mathbb{R}^m$ of a neural network are assigned to log-probabilities $z(x) = \log[p(x)] \in \mathbb{R}^m$ by means of the softmax function $p(x) = \exp(x)/\sum_i \exp(x_i)$. Consequently,

$$z(x) = x - \log\left[\sum_i \exp(x_i)\right],$$

and the Jacobian reads $\mathrm{D}z(x) = I_m - p(x)\mathbf{1}_m^\top$ with $\mathbf{1}_m^\top \in \mathbb{R}^m$ denoting a vector of ones. Moreover, the log-probability Hessians with respect to $x$ are given by $\mathrm{H}z_k(x) = -\mathrm{diag}[p(x)] + p(x)p(x)^\top$.

**Cross-entropy:** The negative log-probabilities are used to compute the cross-entropy with the probability distribution of the label $y \in \mathbb{R}^m$ with $\sum_k y_k = 1$ (usually a one-hot vector),

$$E(z, y) = -y^\top z.$$

Since $E$ is linear in the log-probabilities, that is $\mathrm{H}E(z) = 0$, the HBP is given by

$$\mathcal{H}x = [\mathrm{D}z(x)]^\top \mathrm{H}E(z) [\mathrm{D}z(x)] + \sum_k \mathrm{H}z_k(x) \frac{\partial E(z)}{\partial z_k}$$
$$= \{-\mathrm{diag}[p(x)] + p(x)p(x)^\top\} \sum_k (-y_k)$$
$$= \mathrm{diag}[p(x)] - p(x)p(x)^\top.$$

# D  HBP for convolutional neural networks

The recursive approaches presented in Botev et al. (2017); Chen et al. (2018) tackle the computation of curvature blocks for FCNNs. To the best of our knowledge, an extension to CNNs has not been achieved so far. One reason this might be is that convolutions come with heavy notation that is difficult to deal with in index notation.

Martens and Grosse (2015) provide a procedure for computing a Kronecker-factored approximation of the Fisher for convolutions (KFC). This scheme relies on the property of the Fisher to describe the covariance of the log-likelihood's gradients under the model's distribution. Curvature information is thus encoded in the expectation value, and not by backpropagation of second-order derivatives.

To derive the HBP for convolutions, we use the fact that an efficient implementation of the forward pass is decomposed into multiple operations (see Figure S.4), which can be considered independently by means of our modular approach (see Figure S.5 for details). Our analysis starts by considering the backpropagation of curvature through operations that occur frequently in CNNs. This includes the reshape (Subsection D.1) and extraction operation (Subsection D.2). In Subsection D.3 we outline the modular decomposition and curvature backpropagation of convolution for the two-dimensional case. The approach carries over to other dimensions.

All operations under consideration in this Section are linear. Hence the second terms in Equation (7) vanish. Again, we use the framework of matrix differential calculus (Magnus and Neudecker, 1999) to avoid index notation.

## D.1  Reshape/view

The reshape operation reinterprets a tensor $\mathsf{X} \in \mathbb{R}^{n_1 \times \cdots \times n_x}$ as another tensor $\mathsf{Z} \in \mathbb{Z}^{m_1 \times \cdots \times m_z}$

$$\mathsf{Z}(\mathsf{X}) = \mathrm{reshape}(\mathsf{X}),$$

which possesses the same number of elements, that is $\prod_i n_i = \prod_i m_i$. One example is given by the vec operation. It corresponds to a reshape into a tensor of order one. As the arrangement of elements remains unaffected, $\mathrm{vec}\,\mathsf{Z} = \mathrm{vec}\,\mathsf{X}$, and reshaping corresponds to the identity map on the vectorized input. Consequently, one finds (remember that $\mathcal{H}\mathsf{X}$ is a shorthand notation for $\mathcal{H}\,\mathrm{vec}\,\mathsf{X}$)

$$\mathcal{H}\mathsf{X} = \mathcal{H}\mathsf{Z}.$$

## D.2  Index select/map

Whenever elements of a tensor are selected by an operation, it can be described as a matrix-vector multiplication of a binary matrix $\Pi$ and the vectorized tensor. The mapping is described by an index map $\pi$. Element $j$ of the output $z \in \mathbb{R}^m$ is selected as element $\pi(j)$ from the input $x \in \mathbb{R}^n$. Only elements $\Pi_{j,\pi(j)}$ in the selection matrix $\Pi \in \mathbb{R}^{m \times n}$ are one, while all other entries vanish. Consequently, index selection can be expressed as

$$z_j = x_{\pi(j)} \quad \Leftrightarrow \quad z(x) = \Pi x \quad \text{with} \quad \Pi_{j,\pi(j)} = 1.$$

The HBP is equivalent to the linear layer discussed in Subsection B.1,

$$\mathcal{H}z = \Pi^\top (\mathcal{H}x)\Pi.$$

Applications include max-pooling and the im2col/unfold operation (see Subsection D.3). Average-pooling represents a weighted sum of index selections and can be treated analogously.

## D.3  Convolution

The convolution operation acts on local patches of a multi-channel input of sequences, images, or volumes. In the following, we restrict the discussion to two-dimensional convolution. Figure S.3a illustrates the setting. A collection of filter maps, the *kernel* $\mathsf{W}$, is slid over the spatial coordinates of the input tensor $\mathsf{X}$. In each step, the kernel is contracted with the current area of overlap (the *patch*).

Both the sliding process as well as the structure of the patch area can be controlled by hyperparameters of the operation (kernel size, stride, dilation). Moreover, it is common practice to extend the input tensor, for instance by zero-padding (for an introduction to the arithmetics of convolutions, see Dumoulin and Visin, 2016). The approach presented here is not limited to a certain choice of convolution hyperparameters.

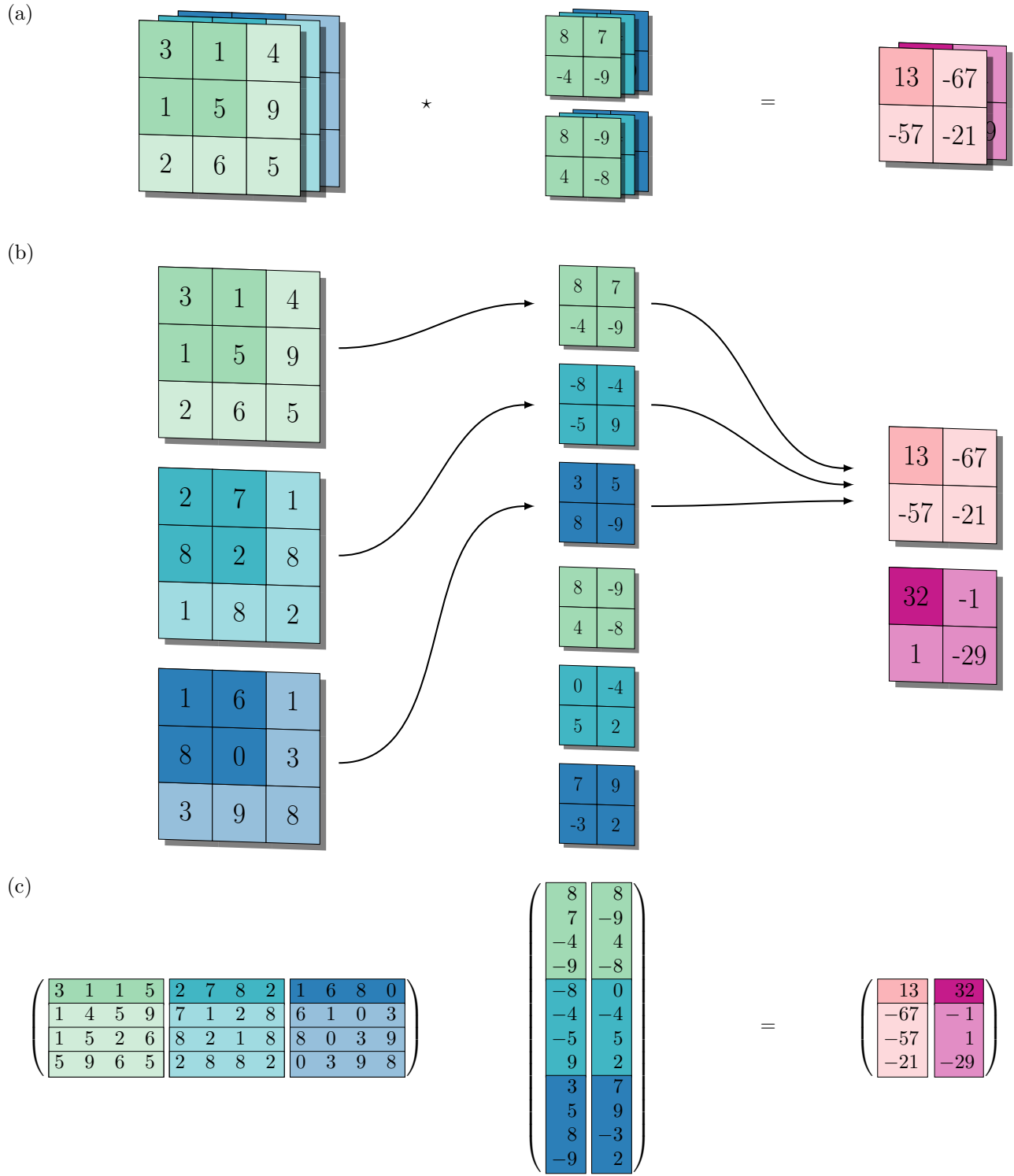

Figure S.3: Two-dimensional convolution $\mathsf{Y} = \mathsf{X} \star \mathsf{W}$ without bias term. (a) The input $\mathsf{X}$ consists of $C_{\text{in}} = 3$ channels (different colors) of $(3 \times 3)$ images. Filter maps of size $(2 \times 2)$ are provided by the kernel $\mathsf{W}$ for the generation of $C_{\text{out}} = 2$ output channels. Patch and kernel volumes that are contracted in the first step of the convolution are highlighted. Assuming no padding and a stride of one results in four patches. New features $\mathsf{Y}$ consist of $C_{\text{out}} = 2$ channels of $(2 \times 2)$ images. (b) Detailed view. All tensors are unrolled along the first axis. (c) Convolution as matrix multiplication. From left to right, the matrices $[\![\mathsf{X}]\!]^\top, W^\top$, and $Y^\top$ are shown.

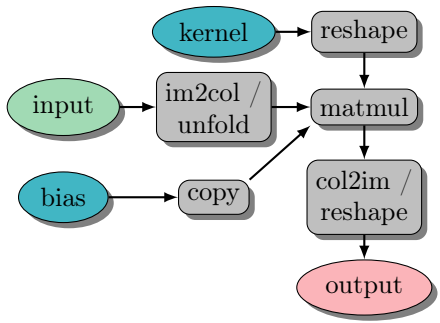

Figure S.4: Decomposition of the convolution operation's forward pass.

Table S.1: Important quantities for the convolution operation. The number of patches equals the product of the outputs' spatial dimensions, i.e. $P = Y_1 Y_2$.

| Tensor | Dimensionality | Description |
| --- | --- | --- |
| $\mathsf{X}$ | $(C_{\text{in}}, X_1, X_2)$ | Input |
| $\mathsf{W}$ | $(C_{\text{out}}, C_{\text{in}}, K_1, K_2)$ | Kernel |
| $\mathsf{Y}$ | $(C_{\text{out}}, Y_1, Y_2)$ | Output |
| $[\![\mathsf{X}]\!]$ | $(C_{\text{in}} K_1 K_2, P)$ | Expanded input |
| $W$ | $(C_{\text{out}}, C_{\text{in}} K_1 K_2)$ | Matricized kernel |
| $Y$ | $(C_{\text{out}}, P)$ | Matricized output |
| $b$ | $C_{\text{out}}$ | Bias vector |
| $B$ | $(C_{\text{out}}, P)$ | Bias matrix |

### D.3.1 Forward pass and notation

We now introduce the quantities involved in the process along with their dimensions. For a summary, see Table S.1. A forward pass of convolution proceeds as follows (cf. Figure S.3b for an example):

- The input $\mathsf{X}$, a tensor of order three, stores a collection of two-dimensional data. Its components $\mathsf{X}_{c_{\text{in}}, x_1, x_2}$ are referenced by indices for the channel $c_{\text{in}}$ and the spatial location $(x_1, x_2)$. $C_{\text{in}}, X_1, X_2$ denote the input channels, width, and height of the image, respectively.

- The kernel $\mathsf{W}$ is a tensor of order four with dimensions $(C_{\text{out}}, C_{\text{in}}, K_1, K_2)$. Kernel width $K_1$ and height $K_2$ determine the patch size $P = K_1 K_2$ for each channel. New features are obtained by contracting the patch and kernel. This is repeated for a collection of $C_{\text{out}}$ output channels stored along the first axis of $\mathsf{W}$.

- Each output channel $c_{\text{out}}$ is shifted by a bias $b_{c_{\text{out}}}$, stored in the $C_{\text{out}}$-dimensional vector $b$.

- The output $\mathsf{Y} = \mathsf{X} \star \mathsf{W}$ with components $Y_{c_{\text{out}}, y_1, y_2}$ is of the same structure as the input. We denote the spatial dimensions of $\mathsf{Y}$ by $Y_1, Y_2$, respectively. Hence $\mathsf{Y}$ is of dimension $(C_{\text{out}}, Y_1, Y_2)$.

**Example (index notation):** A special case where input and output have the same spatial dimensions (Grosse and Martens, 2016) uses a stride of one, kernel widths $K_1 = K_2 = 2K + 1$, $(K \in \mathbb{N})$, and padding $K$. Elements of the filter $\mathsf{W}_{c_{\text{out}}, c_{\text{in}}, :, :}$ are addressed with the index set $\{-K, \ldots, 0, \ldots, K\} \times \{-K, \ldots, 0, \ldots, K\}$. In this setting,

$$\mathsf{Y}_{c_{\text{out}}, y_1, y_2} = \sum_{k_1 = -K}^{K} \sum_{k_2 = -K}^{K} \mathsf{X}_{c_{\text{in}}, x_1 + k_1, x_2 + k_2} \mathsf{W}_{c_{\text{out}}, c_{\text{in}}, k_1, k_2} + b_{c_{\text{out}}} . \tag{S.13}$$

Elements of $\mathsf{X}$ addressed out of bounds evaluate to zero. Arbitrary convolutions come with even heavier notation.

**Convolution by matrix multiplication:** Evaluating convolutions by sums of the form (S.13) leads to poor memory locality (Grosse and Martens, 2016). For improved performance, the computation is mapped to a matrix multiplication (Chellapilla et al., 2006). To do so, patches of the input $\mathsf{X}$ are extracted and flattened into columns of a matrix. The patch extraction is indicated by the operator $[\![\cdot]\!]$ and the resulting matrix $[\![\mathsf{X}]\!]$ is of dimension $(C_{\text{in}} K_1 K_2 \times P)$ (cf. left part of Figure S.3c showing $[\![\mathsf{X}]\!]^\top$). In other words, elements contracted with the kernel are stored along the first axis of $[\![\mathsf{X}]\!]$. $[\![\cdot]\!]$ is also referred to as im2col or unfold operation[1], and accounts for padding.

The kernel tensor $\mathsf{W}$ is reshaped into a $(C_{\text{out}} \times C_{\text{in}} K_1 K_2)$ matrix $W$, and elements of the bias vector $b$ are copied columnwise into a $(C_{\text{out}} \times P)$ matrix $B = b \mathbf{1}_P^\top$, where $\mathbf{1}_P$ is a $P$-dimensional vector of ones. Patchwise contractions are carried out by matrix multiplication and yield a matrix $Y$ of shape $(C_{\text{out}}, P)$ with $P = Y_1 Y_2$,

$$Y = W [\![\mathsf{X}]\!] + B \tag{S.14}$$

(Figure S.3c shows the quantities $W^\top, [\![\mathsf{X}]\!]^\top$, and $Y$ from left to right). Reshaping $Y$ into a $(C_{\text{out}}, Y_1, Y_2)$ tensor, referred to as col2im, yields $\mathsf{Y}$. Figure S.4 summarizes the outlined decomposition of the forward pass.

---
[1] Our definition of the unfold operator slightly differs from Grosse and Martens (2016), where flattened patches are stacked rowwise. This lets us achieve an analogous form to a linear layer. Conversion is achieved by transposition.

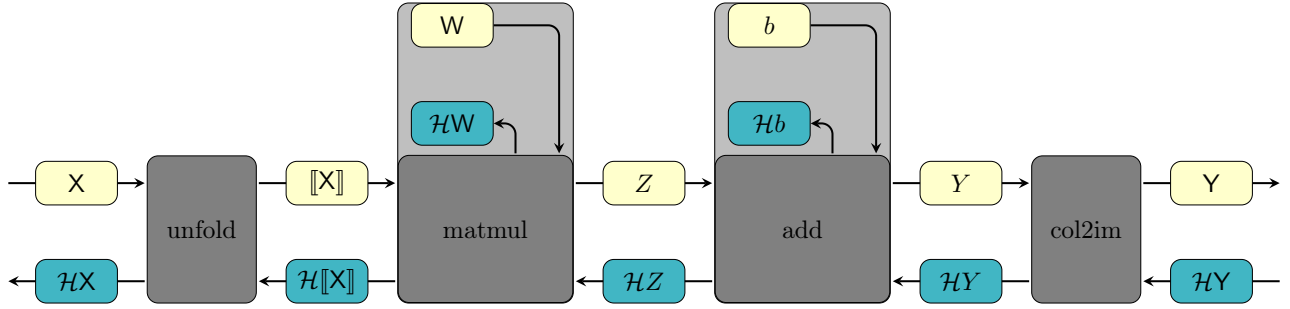

Figure S.5: Decomposition of convolution with notation for the study of curvature backpropagation.

### D.4 HBP for convolution

We now compose the HBP for convolution, proceeding from right to left with the operations depicted in Figure S.5, by analyzing the backpropagation of curvature for each module, adopting the figure's notation.

**Col2im/reshape:** The col2im operation takes a matrix $Y \in \mathbb{R}^{C_\text{out} \times Y_1 Y_2}$ and reshapes it into the tensor $\mathsf{Y} \in \mathbb{R}^{C_\text{out} \times Y_1 \times Y_2}$. According to Subsection D.1, $\mathcal{H}Y = \mathcal{H}\mathsf{Y}$.

**Bias Hessian:** Forward pass $Y = Z + B$ and Equation (S.8b) imply $\mathcal{H}Y = \mathcal{H}B = \mathcal{H}Z$. To obtain the Hessian with respect to $b$ from $\mathcal{H}B$, consider the columnwise copy operation $B(b) = b\mathbf{1}_P^\top$, whose matrix differential is $\mathrm{d}B(b) = (\mathrm{d}b)\mathbf{1}_P^\top$. Vectorization yields $\mathrm{d}\,\mathrm{vec}\,B(b) = (\mathbf{1}_P \otimes I_{C_\text{out}})\mathrm{d}b$. Hence, the Jacobian is $\mathrm{D}B(b) = \mathbf{1}_P \otimes I_{C_\text{out}}$, and insertion into Equation (7) results in

$$\mathcal{H}b = (\mathbf{1}_P \otimes I_{C_\text{out}})^\top \mathcal{H}B\, (\mathbf{1}_P \otimes I_{C_\text{out}})\,.$$

This performs a linewise and columnwise summation over $\mathcal{H}B$, summing entities that correspond to copies of the same entry of $b$ in the matrix $B$. It can also be regarded as a reshape of $\mathcal{H}B$ into a $(C_\text{out}, P, C_\text{out}, P)$ tensor, which is then contracted over the second and fourth axis.

**Weight Hessian:** For the matrix-matrix multiplication $Z(W, [\![\mathsf{X}]\!]) = W[\![\mathsf{X}]\!]$, the HBP was discussed in Subsection B.1. The Jacobians are given by $\mathrm{D}Z([\![\mathsf{X}]\!]) = I_P \otimes W$ and $\mathrm{D}Z(W) = [\![\mathsf{X}]\!]^\top \otimes I_S$ with the patch size $S = C_\text{in}K_1K_2$. Hence, the HBP for the unfolded input and the weight matrix are

$$\mathcal{H}[\![\mathsf{X}]\!] = (I_P \otimes W)^\top \mathcal{H}Z\,(I_P \otimes W)\,,$$
$$\mathcal{H}W = ([\![\mathsf{X}]\!]^\top \otimes I_S)^\top \mathcal{H}Z\,([\![\mathsf{X}]\!]^\top \otimes I_S)\,.$$

From what has been said about the reshape operation in Subsection D.1, it follows that $\mathcal{H}\mathsf{W} = \mathcal{H}W$.

**Im2col/unfold:** The patch extraction operation $[\![\cdot]\!]$ copies all elements in a patch into the columns of a matrix and thus represents a selection of elements by an index map, which is hard to express in notation. Numerically, it can be obtained by calling im2col on a $(C_\text{in}, X_1, X_2)$ index tensor whose entries correspond to the indices. The resulting tensor contains all information about the index map. HBP follows the relation of Subsection D.2.

**Discussion:** Although convolution can be understood as a matrix multiplication, the parameter Hessian is not identical to that of the linear layer discussed in Subsection B.1. The difference is due to the parameter sharing of the convolution. In the case of a linear layer $z = Wx + b$, the Hessian of the weight matrix for a single sample possesses Kronecker structure (Botev et al., 2017; Chen et al., 2018; Bakker et al., 2018), i.e. $\mathcal{H}W = x \otimes x^\top \otimes \mathcal{H}z$. For convolution layers, however, it has been argued by Bakker et al. (2018) that block diagonals of the Hessian do not inherit the same structure. Rephrasing the forward pass (S.14) in terms of vectorized quantities, we find

$$\mathrm{vec}\,Y = (I_P \otimes W)\,\mathrm{vec}[\![\mathsf{X}]\!] + \mathrm{vec}\,B\,.$$

In this perspective, convolution corresponds to a fully-connected linear layer, with the additional constraints that the weight and bias matrix be circular. Defining $\hat{W} = I_P \otimes W$, one then finds the Hessian $\mathcal{H}\hat{W}$ to possess Kronecker structure. Parameterization with a kernel tensor encodes the circularity constraint in weight sharing.

Table S.2: Model architectures under consideration. We use the patterns Linear(in_features, out_features), Conv2d(in_channels, out_channels, kernel_size, padding), MaxPool2d(kernel_size, stride), and ZeroPad2d(padding_left, padding_right, padding_top, padding_bottom) to describe module hyperparameters. Convolution strides are always chosen to be one. (a) FCNN used to extend the experiment in Chen et al. (2018) (3 846 810 parameters). (b) CNN architecture (1 099 226 parameters). (c) DeepOBS c3d3 test problem with three convolutional and three dense layers (895 210 parameters). ReLU activation functions are replaced by sigmoids.

(a) FCNN (Figure 4)

| # | Module |
|---|---|
| 1 | Flatten |
| 2 | Linear(3072, 1024) |
| 3 | Sigmoid |
| 4 | Linear(1024, 512) |
| 5 | Sigmoid |
| 6 | Linear(512, 256) |
| 7 | Sigmoid |
| 8 | Linear(256, 128) |
| 9 | Sigmoid |
| 10 | Linear(128, 64) |
| 11 | Sigmoid |
| 12 | Linear(64, 32) |
| 13 | Sigmoid |
| 14 | Linear(32, 16) |
| 15 | Sigmoid |
| 16 | Linear(16, 10) |

(b) CNN (Figure 5)

| # | Module |
|---|---|
| 1 | Conv2d(3, 16, 3, 1) |
| 2 | Sigmoid |
| 3 | Conv2d(16, 16, 3, 1) |
| 4 | Sigmoid |
| 5 | MaxPool2d(2, 2) |
| 6 | Conv2d(16, 32, 3, 1) |
| 7 | Sigmoid |
| 8 | Conv2d(32, 32, 3, 1) |
| 9 | Sigmoid |
| 10 | MaxPool2d(2, 2) |
| 11 | Flatten |
| 12 | Linear(2048, 512) |
| 13 | Sigmoid |
| 14 | Linear(512, 64) |
| 15 | Sigmoid |
| 16 | Linear(64, 10) |

(c) DeepOBS c3d3 (Figure S.6)

| # | Module |
|---|---|
| 1 | Conv2d(3, 64, 5, 0) |
| 2 | Sigmoid |
| 3 | ZeroPad2d(0, 1, 0, 1) |
| 4 | MaxPool2d(3, 2) |
| 5 | Conv2d(64, 96, 3, 0) |
| 6 | Sigmoid |
| 7 | ZeroPad2d(0, 1, 0, 1) |
| 8 | MaxPool2d(3, 2) |
| 9 | ZeroPad2d(1, 1, 1, 1) |
| 10 | Conv2d(96, 128, 3, 0) |
| 11 | Sigmoid |
| 12 | ZeroPad2d(0, 1, 0, 1) |
| 13 | MaxPool2d(3, 2) |
| 14 | Flatten |
| 15 | Linear(1152, 512) |
| 16 | Sigmoid |
| 17 | Linear(512, 256) |
| 18 | Sigmoid |
| 19 | Linear(256, 10) |

For the Hessian $\mathcal{H}W$ of the kernel to possess Kronecker structure, the output Hessian $\mathcal{H}Z$ has to be assumed to factorize into a Kronecker product of $S \times S$ and $C_{\text{out}} \times C_{\text{out}}$ matrices. These assumptions are somewhat in parallel with the additional approximations introduced in Grosse and Martens (2016) to obtain KFC.

## E  Experimental details

**Fully-connected neural network:** The same model as in Chen et al. (2018) (see Table S.2a) is used to extend the experiment performed therein. The weights of each linear layer are initialized with the Xavier method of Glorot and Bengio (2010). Bias terms are intialized to zero. Backpropagation of the Hessian uses approximation (9) of (S.11) to compute the curvature blocks of the weights and bias, $\overline{\mathcal{H}W^{(i)}}$ and $\overline{\mathcal{H}b^{(i)}}$.

Hyperparameters are chosen as follows to obtain consistent results with the original work: All runs shown in Figure 4 use a batch size of $|B| = 500$. For SGD, the learning rate is assigned to $\gamma = 0.1$ with momentum $v = 0.9$. Block-splitting experiments with the second-order method use the PCH-abs. All runs were performed with a learning rate $\gamma = 0.1$ and a regularization strength of $\alpha = 0.02$. For the convergence criterion of CG, the maximum number of iterations is restricted to $n_{\text{CG}} = 50$; convergence is reached at a relative tolerance $\epsilon_{\text{CG}} = 0.1$.

**Convolutional neural net:** The training procedure of the CNN architecture shown in Table S.2b is evaluated on a hyperparameter grid. Runs with smallest final training loss are selected to rerun on different random seeds. The curves shown in Figure 5b represent mean values and standard deviations for ten different realizations over the random seed. All layer parameters were initialized with the default method in PyTorch.

For the first-order optimizers (SGD, Adam), we considered batch sizes $B \in \{100, 200, 500\}$. In the case of SGD, momentum $v$ was tuned over the set $\{0, 0.45, 0.9\}$. Although we varied the learning rate over a large range of values $\gamma \in \{10^{-3}, 10^{-2}, 0.1, 1, 10\}$, losses kept plateauing and did not decrease. In particular, the loss function even

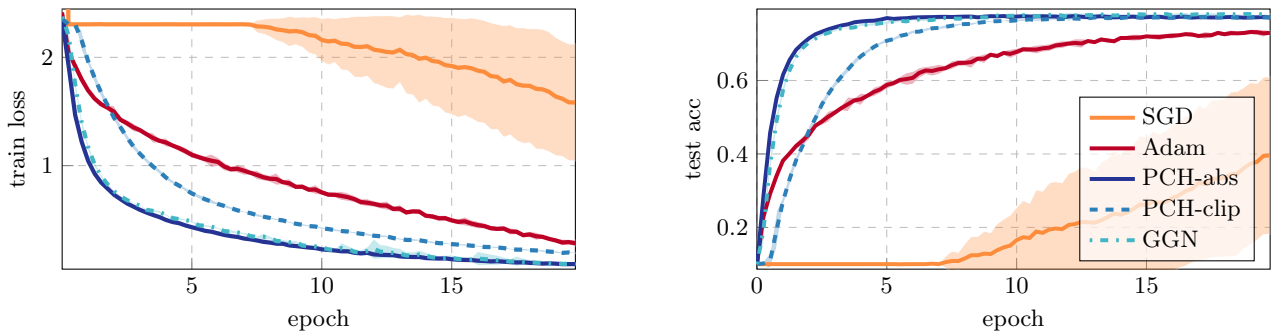

Figure S.6: Comparison of SGD, Adam, and Newton-style methods with different exact curvature matrix-vector products (HBP) on the DeepOBS c3d3 network (Schneider et al., 2019) with sigmoid activations (Table S.2c).

increased for the large learning rates. For Adam, we only vary the learning rate $\gamma \in \{10^{-4}, 10^{-3}, 10^{-2}, 0.1, 1, 10\}$.

As second-order methods scale better to large batch sizes, we considered $B \in \{200, 500, 1000\}$ for them. The convergence parameters for CG were fixed to $n_{\text{CG}} = 200$ and $\epsilon_{\text{CG}} = 0.1$. For all curvature matrices, we varied the learning rates over the grid $\gamma \in \{0.05, 0.1, 0.2\}$ and $\alpha \in \{10^{-4}, 10^{-3}, 10^{-2}\}$.

In order to compare with another second-order method, we experimented with a public PyTorch implementation[2] of the KFAC optimizer (Martens and Grosse, 2015; Grosse and Martens, 2016). All hyperparameters were kept at their default setting. The learning rate was varied over $\gamma \in \{10^{-4}, 10^{-3}, 10^{-2}, 0.1, 1, 10\}$.

The hyperparameters of results shown in Figure 5 read as follows:

- SGD ($|B| = 100, \gamma = 10^{-3}, v = 0.9$). The particular choice of these hyperparameters is artificial. This run is representative for SGD, which does not achieve any progress at all.

- Adam ($|B| = 100, \gamma = 10^{-3}$)

- KFAC ($|B| = 500, \gamma = 0.1$)

- PCH-abs ($|B| = 1000, \gamma = 0.2, \alpha = 10^{-3}$), PCH-clip ($|B| = 1000, \gamma = 0.1, \alpha = 10^{-4}$)

- GGN, $\alpha_1$ ($|B| = 1000, \gamma = 0.1, \alpha = 10^{-4}$). This run does not yield the minimum training loss on the grid, and is shown to illustrate that the second-order method is capable to escape the flat regions in early stages.

- GGN, $\alpha_2$ ($|B| = 1000, \gamma = 0.1, \alpha = 10^{-3}$). In comparison with $\alpha_1$, the second-order method requires more iterations to escape the initial plateau, caused by the larger regularization strength. However, this leads to improved robustness against noise in later stages of the training procedure.

**Additional experiment:** Another experiment conducted with HBP considers the c3d3 architecture (Table S.2c) of DeepOBS (Schneider et al., 2019) on CIFAR-10 . ReLU activations are replaced by sigmoids to make the problem more challenging. The hyperparameter grid is chosen identically to the CNN experiment above, and results are summarized in Figure S.6. In particular, the hyperparameter settings for each competitor are:

- SGD ($|B| = 100, \gamma = 1, v = 0$)

- Adam ($|B| = 100, \gamma = 10^{-3}$)

- PCH-abs ($|B| = 500, \gamma = 0.1, \alpha = 10^{-3}$), PCH-clip ($|B| = 500, \gamma = 0.1, \alpha = 10^{-2}$)

- GGN ($|B| = 500, \gamma = 0.05, \alpha = 10^{-3}$)

---

[2] https://github.com/alecwangcq/KFAC-Pytorch



\end{document}